\documentclass[twocolumn]{article}

\usepackage[utf8]{inputenc}
\usepackage{titling}
\usepackage{color}
\usepackage{float}
\usepackage{units}
\usepackage{amsmath}
\usepackage{amssymb}
\usepackage{graphicx}
\usepackage{color}
\usepackage{tikz}
\usepackage{subfig}
\usepackage{xcolor}
\usepackage{algorithm}
\usepackage{algpseudocode}
\usepackage{soul}
\usepackage{authblk}

\makeatletter

\newtheorem{theorem}{Theorem}

\newtheorem{remark}[theorem]{Remark}

\let\oldReturn\Return
\renewcommand{\Return}{\State\oldReturn}

\@ifundefined{showcaptionsetup}{}{%
	\PassOptionsToPackage{caption=false}{subfig}}
\usepackage{subfig}
\makeatother

\usetikzlibrary{arrows}
\usetikzlibrary{decorations.pathreplacing}
\usetikzlibrary{fit,calc,positioning,decorations.pathreplacing,matrix,patterns}

\def\:{\colon}

\newcommand{\R}{\mathbb{R}}

\renewcommand{\deg}{\mathrm{deg\,}}

\newcommand{\actual}[1]{#1^{*}}
\newcommand{\lb}[1]{#1^-}
\newcommand{\ub}[1]{#1^+}

\newcommand{\Long}[1]{}
\usepackage{ifthen}

\makeatletter
\newcommand\footnoteref[1]{\protected@xdef\@thefnmark{\ref{#1}}\@footnotemark}
\makeatother

\usepackage{moreverb,url}

\usepackage[colorlinks,bookmarksopen,bookmarksnumbered,citecolor=red,urlcolor=red]{hyperref}

\providecommand{\keywords}[1]{\textbf{\textit{Index terms---}} #1}

\begin{document}
\title{Proving the existence of loops\\in robot trajectories}

\author[1]{Simon Rohou}
\author[2]{Peter Franek}
\author[3]{Cl\'{e}ment Aubry}
\author[1]{Luc Jaulin}
\affil[1]{\small ENSTA Bretagne, Lab-STICC, UMR CNRS 6285, Brest, France}
\affil[2]{\small IST Austria, Am Campus 1, 3400 Klosterneuburg, Austria}
\affil[3]{\small ISEN Brest, L@bISEN, France}

\date{}

\maketitle
\begin{abstract}
	This paper presents a~reliable method to verify the existence of loops along the uncertain trajectory of a robot, based on proprioceptive measurements only, within a bounded-error context. The loop closure detection is one of the key points in SLAM methods, especially in homogeneous environments with difficult scenes recognitions. The proposed approach is generic and could be coupled with conventional SLAM algorithms to reliably reduce their computing burden, thus improving the localization and mapping processes in the most challenging environments such as unexplored underwater extents.
	To prove that a~robot performed a loop whatever the uncertainties in its evolution, we employ the notion of \emph{topological degree} that originates in the field of differential topology. We show that a~verification tool based on the topological degree is an~optimal method for proving robot loops. This is demonstrated both on datasets from real missions involving autonomous underwater vehicles, and by a~mathematical discussion.
	
	
\end{abstract}

\section{Introduction}
\label{sec:introduction}

	The SLAM, Simultaneous Localization And Mapping \cite{SlamOrigins,bosse2004}, is an approach that ties together the problem of state estimation and the one of mapping an unknown environment. Basically, a robot coming back to a previous pose is likely to recognize an old scene and then refine its localization. The key point of these methods is then to detect that a place has been previously visited. This problem of data association is known in the literature as loop closure \cite{latif2013}.

	\subsection{Detecting loop closures}
		
		A loop can be detected thanks to \emph{exteroceptive} measurements, \emph{i.e.} the perception of the outside, by scenes comparisons \cite{Ange:Fast:08,cummins2008fab,stachniss2004exploration,clemente2007mapping}.
		However, it can be difficult to detect the closure due to poor estimations on both the robot's position and map-matchings. The problem appears even more challenging when dealing with homogeneous environments without any point of interest to rely on. This is typically the case one can encounter in underwater exploration with wide homogeneous sea-floors. Such situation will unfortunately lead to a few detections of confident loop closures or, in the worst cases, to false detections that could lead to a wrong localization and mapping.
		
		Besides exteroceptive measurements, it has been shown in \cite{aubry:autom:13} that loops can be approximated based on \emph{proprioceptive} measurements only, namely: velocity vectors and inertial values knowing the kinematic of the robot.
		This approach has the advantage to be applicable regardless of the nature of the environment to explore.
		Of course, one should note that in this very case, the loop detections cannot improve by themselves the localization, as the approach will not bring new information or constraints to the problem.
		
		However, this method is of high interest if combined with classical SLAM techniques that merge both proprioceptive and exteroceptive measurements, in order to decrease the computing burden of usual scenes recognitions. Indeed, the complexity of SLAM algorithms quickly increases with the exploration of wide environments, as it implies lots of loop closures to identify among a dense set of data. To this day, the execution of SLAM programs in 3D environments during long-term missions is often not affordable for classical embedded systems powering the robots. A part of the community hence focuses on lighter and embeddable solutions. This work is heading in this direction, proposing a way to estimate the loop closures that does not rely on environment observations. This approach is then guaranteed to provide real-time results as it does not go into a costly analysis of heavy observation datasets.
		
\begin{figure}[!h]
			\centering
			\includegraphics[width=1.0\linewidth]{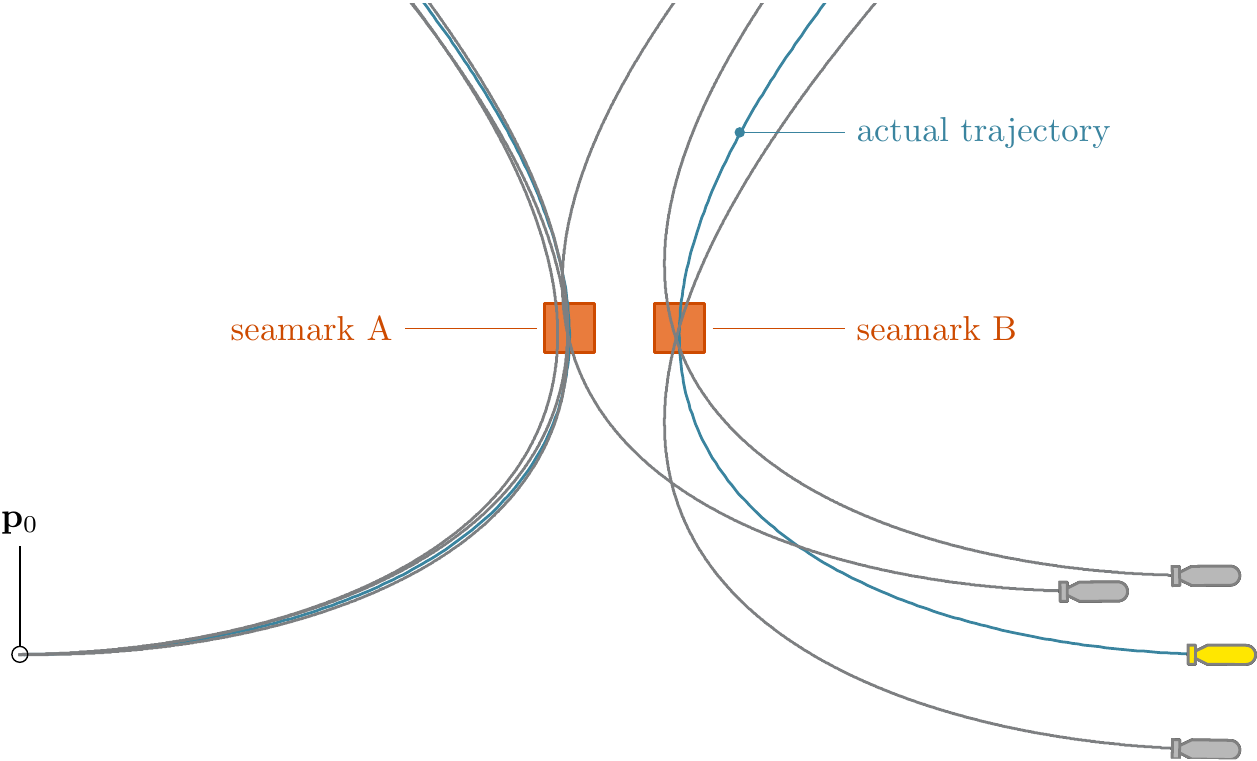}
			\caption{A robot flying over two different but same-looking seamarks. The actual trajectory is plotted in blue while several dead-reckoning estimations are drawn in gray. All the trajectories are consistent with the observations. A well-known map would not prevent from wrong loop closure detections.}
			\label{fig:false:loop:closure}
		\end{figure}

		On top of that, a reliable approach that provides guaranteed loop approximations is suited to prevent from false detections in singular environments. This situation is typically encountered when two different objects of same shape are considered as unique by algorithms standing on too uncertain positioning estimations. Figure~\ref{fig:false:loop:closure} gives an example of same looking objects and uncertain trajectories estimations. This situation may lead to the detection of wrong loop closures.
		Our method provides a way to reject the feasibility of a loop closure despite the ambiguity of the situation.
		  
	\subsection{The two-dimensional case}
		
		Formally, a robot that performed a loop is a robot that came back to a previous position $\mathbf{p}(t)$.
		We will focus on the detection of loops along two-dimensional trajectories: $\mathbf{p}(t)\in\mathbb{R}^2$. This choice is not a limitation made to keep things simple, but a practical requirement. Indeed, it is not possible to physically verify $\mathbf{p}(t_1)=\mathbf{p}(t_2)$ in higher dimensional spaces. A robot will never reach again the very same 3D atomic position, in contrast with two-dimensional cases. Furthermore, the amount of uncertainties we have to deal with will always be too large to verify this. Therefore, it is not possible to prove three-dimensional loops, nor to verify that a robot came back to a previous \emph{pose}, including both position and orientation, for the same reason.
		
\begin{figure}[!h]
			\centering
			\includegraphics[width=1.0\linewidth]{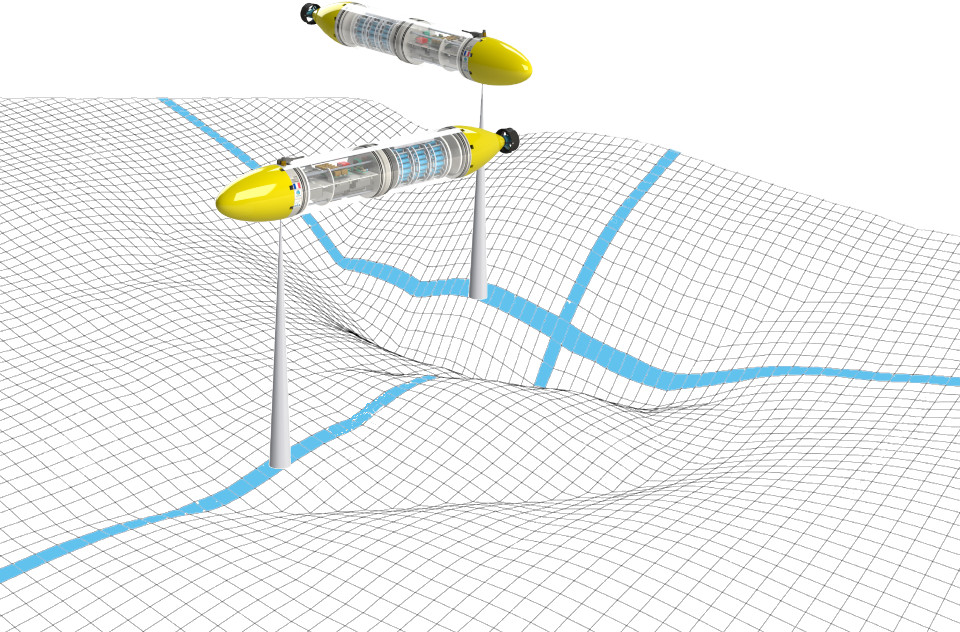}
			\caption{An underwater robot exploring its environment with a single beam echo-sounder. This view presents two instants of the mission, before and after performing a loop. The robot trajectory is projected in blue on the sea-floor.}
			\label{fig:loop:3d:view}
		\end{figure}

		Verify a two-dimensional loop is still of interest for many 3D applications. For instance, as pictured in Figure~\ref{fig:loop:3d:view}, an underwater robot can apply a raw-data SLAM method using a sonar for exteroceptive measurements. In this configuration, the SLAM can be reduced to a 2D problem by merging vertical measurements, namely: depth from a pressure sensor and altitude from the sonar. Map-matching will then be achievable over each 2D crossing, as pictured in the figure with projections on the sea-floor.
		
	\vfill
	
	The main contribution of this paper is to provide a reliable \emph{existence test} that will verify a~given loop closure detection. Such test has already been the subject of \cite{aubry:autom:13} with a proposition based on the Newton operator \cite{Moore79}. However, this test $\mathcal{N}$ is not always able to conclude on obvious existence cases, as it stands on a Jacobian matrix that is sometimes not invertible. Our contribution is to propose a new test $\mathcal{T}$ relying on the topological degree theory \cite{cho2006topological,franek2012} that outperforms the previous method, thus increasing the number of proofs of loop closures on robot trajectories.
	
	This paper is organized as follows. Section \ref{sec:loops} details how loops can be detected thanks to proprioceptive measurements, especially in a bounded-error context. It is shown that proving the existence of a loop amounts to checking that an uncertain function vanishes at some point, which can be verified thanks to the topological degree theory presented in Section \ref{sec:topodegree}. This theoretical part applied on our loop problem is implemented under a new dedicated existence test provided in Section \ref{sec:test:existence}. The same tool is extended in Section \ref{sec:test:uniqueness} for uniqueness verification purposes in order to prove that a given detection set encloses a unique solution for a loop. The proposed algorithms are then applied on an actual experiment described in Section \ref{sec:application}, before a discussion about the optimality of the method in Section \ref{sec:optimal} and the conclusion of the paper.

\section{Proprioceptive\\loop detections}
\label{sec:loops}

	This section details how loops can be detected thanks to proprioceptive measurements only.
	We recall that \emph{proprioceptive} measurements shall mean values about robot's states sensed by the robot itself, for instance: velocity, inertial values, heading, \emph{etc}. A definition of a loop set is provided, before details about guaranteed tools that will be used then for loop detections in a bounded-error context.
	
	\subsection{Formalization}
	\label{sec:loops:formalization}
	
\begin{figure}[!h]
			\centering
			\includegraphics[width=\linewidth]{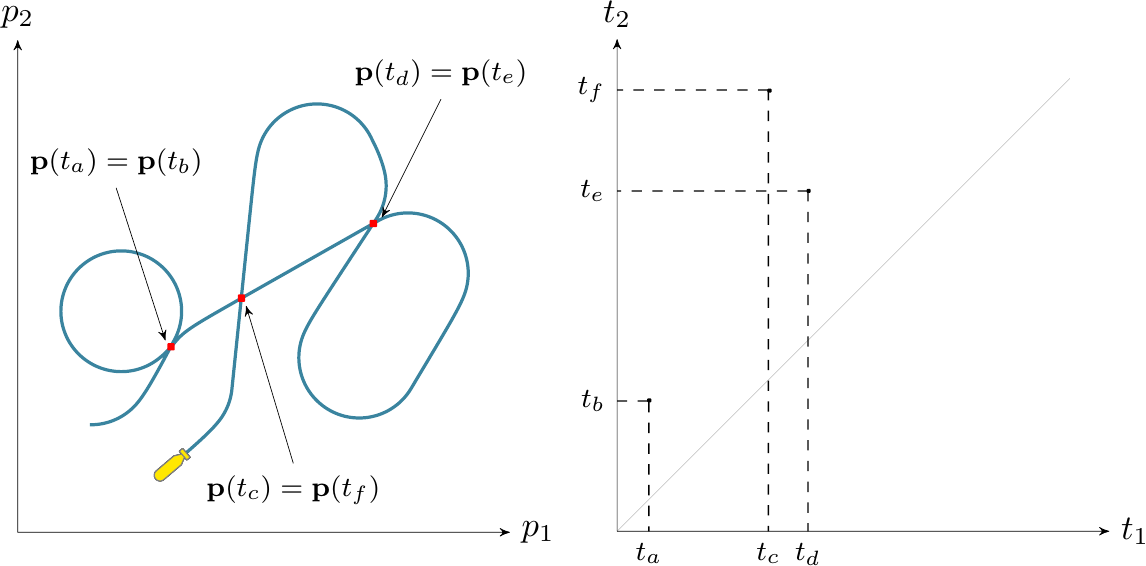}
			\caption{A robot performing three loops: its own trajectory has been crossed three times. A temporal representation provided by the $t$-plane (right-hand side) is used to picture the loops by $t$-pairs $(t_a,t_b)$, $(t_c,t_f)$, $(t_d,t_e)$.}
			\label{fig:tplane:loops:def}
		\end{figure}

		In \cite{aubry:autom:13}, a loop is defined by a $t$-pair $(t_1,t_2)$ such that $\mathbf{p}(t_1)=\mathbf{p}(t_2)$, $t_1\not=t_2$, where $\mathbf{p}(t)$ is the two-dimensional position of the robot at $t$. The loop detection consists in computing the set $\actual{\mathbb{T}}$ of all loops:
		\begin{equation}
			\actual{\mathbb{T}}=\left\{(t_1,t_2)\in [t_0,t_f]^2 \mid \mathbf{p}(t_1)=\mathbf{p}(t_2) , t_1 < t_2 \right\},
			\label{eq:T:star}
		\end{equation} with $t_0, t_f$ being respectively the start and the end times of a~trajectory.
		Graphically, we represent the \emph{loop set} $\actual{\mathbb{T}}$ as a~set of points in the $t$-plane.
		An example of $\actual{\mathbb{T}}=\left\{(t_a,t_b), (t_c,t_f), (t_d,t_e)\right\}$ is provided in Figure~\ref{fig:tplane:loops:def}.
		
		We consider a mobile robot moving on a horizontal plane.
		Its trajectory is made of several 2D~positions defined by
		\begin{equation}
			\mathbf{p}(t)=\int_{t_0}^t \mathbf{v}(\tau)d\tau + \mathbf{p}_0,
		\end{equation}
		where $\mathbf{v}(t)\in\mathbb{R}^2$ is the velocity vector of the robot at time $t\in[t_0,t_f]$ expressed in the environment reference frame. $\mathbf{v}(t)$ is a \emph{proprioceptive} information that can be easily sensed by the robot at any time. Then, the loop set $\actual{\mathbb{T}}$ is
		\begin{equation}
			\actual{\mathbb{T}}=\left\{(t_1,t_2)\in [t_0,t_f]^2 \mid \int_{t_1}^{t_2} \mathbf{v}(\tau)d\tau = \mathbf{0} , t_1 < t_2 \right\},
			\label{eq:T:star:velocities}
		\end{equation}
		which means that for any $(t_1,t_2)\in\actual{\mathbb{T}}$, robot's move from $t_1$ vanishes at $t_2$. Therefore, any loop can be detected based on these velocity measurements.
		
		In practice, these values are noisy and we assume the measurements are performed with a known bounded error \cite{Meizel96}, \emph{i.e.} a box $[\mathbf{v}](t)$ contains the actual $\actual{\mathbf{v}}(t)$ for each $t\in[t_0,t_f]$. This set-membership approach will stand on interval analysis, a mathematical field that appeared during the last decades \cite{Moore66} and  is particularly suitable for verified computing.
		This tool is briefly presented hereinafter.
		
	\subsection{Tools for\\guaranteed computations}
	
		This section first introduces basic notions of interval analysis \cite{Moore79,Hansen65} before focusing on tubes that will be used to handle proprioceptive measurements and their uncertainties over time.
		
		\subsubsection{Interval analysis\\}
		
			An interval $[x]=[\lb{x},\ub{x}]=\left\{x\in\mathbb{R} \mid \lb{x}\leqslant x \leqslant \ub{x}\right\}$ is a closed and connected subset of $\mathbb{R}$ delimited by a lower bound $\lb{x}$ and an upper one $\ub{x}$. A Cartesian product of $n$ intervals defines a \emph{box} -- also called \emph{interval-vector} -- belonging to the set $\mathbb{IR}^n$. In this paper, intervals are written into brackets and vectors and boxes are represented in bold: $[\mathbf{x}]$. The actual but unknown value, enclosed within a box, is denoted by a star: $\actual{\mathbf{x}}$.
			
			Interval analysis is based on the extension of all classical real arithmetic operators $+$, $-$, $\times$ and $\div$. For instance:
			$$[x]+[y]=[\lb{x}+\lb{y},\ub{x}+\ub{y}],$$
			$$[x]-[y]=[\lb{x}-\ub{y},\ub{x}-\lb{y}].$$
			This extension also includes the adaptation of elementary functions such as $\cos$, $\exp$, $\tan$. The output is the smallest interval containing all the images of all defined inputs through the function.

		\subsubsection{Tubes\\}\label{sst:tubes}
		
\begin{figure}[h]
				\centering
				\includegraphics[width=\linewidth]{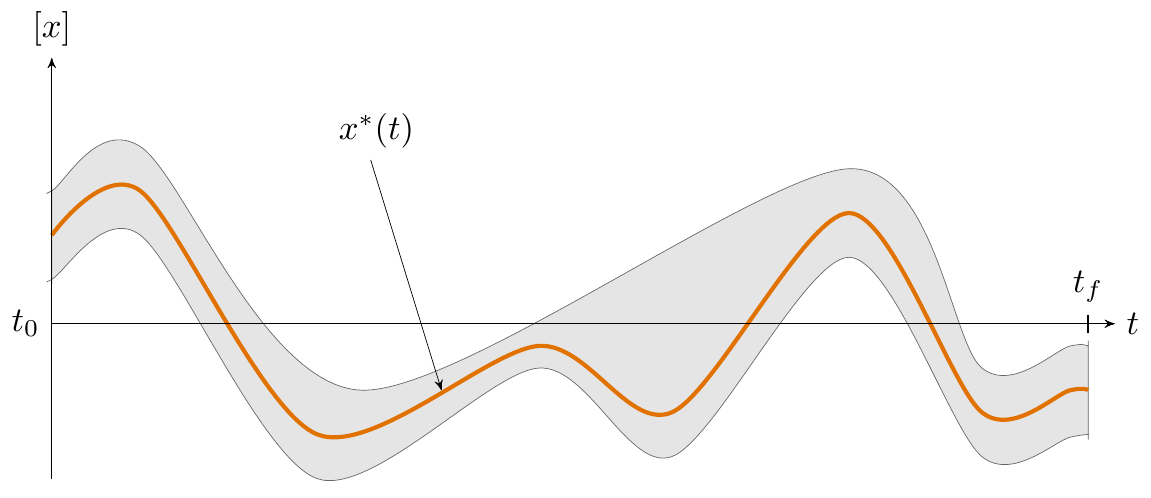}
				\caption{A tube $[x](t)$ with domain $[t_0,t_f]$ enclosing an unknown trajectory $\actual{x}(t)$. The thinner the tube, the better the approximation of $\actual{x}(t)$.}
				\label{fig:tube:def}
			\end{figure}

			Classical intervals of reals can be extended to trajectories by means of \emph{tubes}.		
			A tube \cite{Kurzhanski97,LebarsTubes12} $[\mathbf{x}](t):\mathbb{R}\to\mathbb{IR}^n$ is an envelope enclosing an uncertain trajectory denoted by $\actual{\mathbf{x}}(t):\mathbb{R}\to\mathbb{R}^n$. This enclosure can be defined as an interval of two functions $\lb{\mathbf{x}}(t)$ and $\ub{\mathbf{x}}(t)$ such that $\forall t\in[t_0,t_f] , \lb{\mathbf{x}}(t) \leqslant \ub{\mathbf{x}}(t)$. Figure~\ref{fig:tube:def} gives an illustration of a tube enclosing a trajectory $\actual{x}(t):\mathbb{R}\to\mathbb{R}$.
			As for intervals, tubes can be handled with the extension of classical real arithmetic operators (such as addition, usual functions, \emph{etc.}). This can be done using interval arithmetic applied on each $t$ of tube's domain.
			
\begin{figure}[h]
				\centering
				\includegraphics[width=\linewidth]{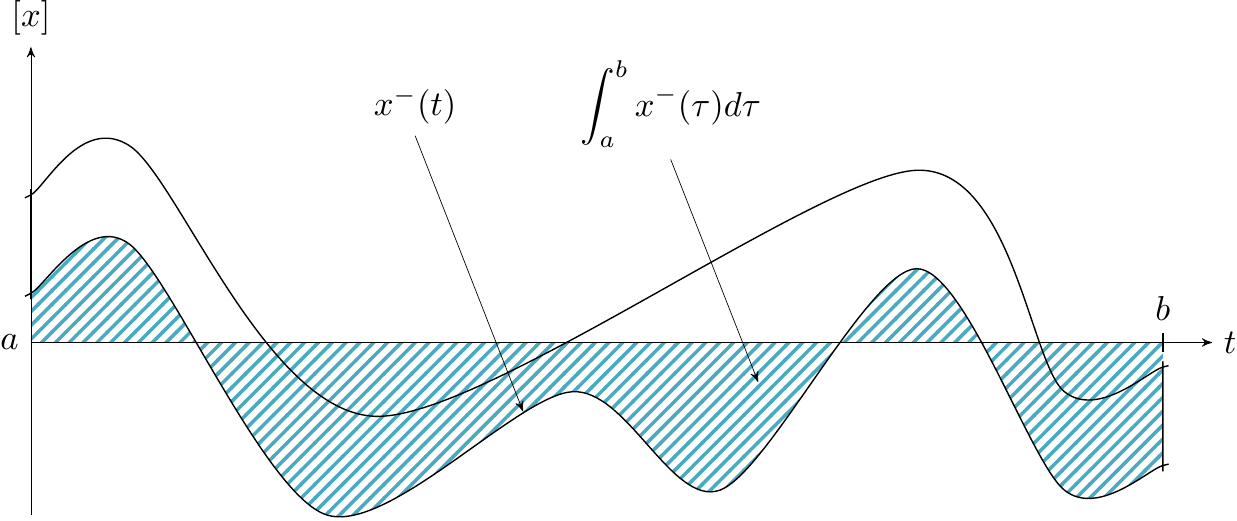}
				\caption{Lower bound of the integral of a tube. Hatched part depicts the lower bound of $\int_{a}^{b}[x](\tau)d\tau$.}
				\label{fig:tube:integ}
			\end{figure}

			The integral of a tube is defined from $t_1$ to $t_2$ as the smallest box containing all feasible integrals:
			\begin{equation}
				\int_{t_1}^{t_2}[\mathbf{x}](\tau)d\tau=\biggl\{\int_{t_1}^{t_2}\mathbf{x}(\tau)d\tau\mid\mathbf{x}(\cdot)\in[\mathbf{\mathbf{x}}](\cdot)\biggr\}.
			\end{equation}
			From the monotonicity of the integral operator, we can deduce:
			\begin{equation}
				\int_{t_1}^{t_2}[\mathbf{x}](\tau)d\tau=\biggr[\int_{t_1}^{t_2}\lb{\mathbf{x}}(\tau)d\tau,\int_{t_1}^{t_2}\ub{\mathbf{x}}(\tau)d\tau\biggl].
			\end{equation}
			The lower bound of this box is illustrated by Figure~\ref{fig:tube:integ}. The integral can also be computed between bounded bounds $[t_1]$, $[t_2]$ by
			\begin{eqnarray}
				\begin{array}{rcl}
				\int_{[t_1]}^{[t_2]}[\mathbf{x}](\tau)d\tau & = & \bigr[\textrm{lb}\left(\lb{\mathbf{y}}([t_2])-\lb{\mathbf{y}}([t_1])\right),\\
				~ & ~ & ~\textrm{ub}\left(\ub{\mathbf{y}}([t_2])-\ub{\mathbf{y}}([t_1])\right)\bigl]
				\end{array},
				\label{eq:tubes:boundedbounds}
			\end{eqnarray}
			where $[\mathbf{y}](t)=\int_{t_0}^{t}[\mathbf{x}](\tau)d\tau$ is the interval primitive of $[\mathbf{x}](\cdot)$ and $\mathbf{y}^{\pm}$ are the corresponding bounds.
			The proof is provided in \cite[Sec. 3.3]{aubry:autom:13}.
			
			A tube is generally used to describe uncertain trajectories evolving with time and defined by differential equations \cite{RohouTubInt,RaissiAutomatica04,GoldsztejnHayes11}. This is naturally of high interest in robotics, being useful for dynamical systems such as mobile robots, involving uncertainties and any kind of temporal constraints.
	
	\subsection{Loop detections in a bounded-error context}
		
		It has been shown in Section \ref{sec:loops:formalization} that a loop can be detected based on velocity measurements.
		In practice, trajectories are estimated by measurements corrupted by noise, leading to spatial uncertainties. Hence, from Eq. \eqref{eq:T:star:velocities}, the set of $t$-pairs cannot be computed exactly. In a set-membership context \cite{drevelleTRO13,GningJM13}, measurement errors are bounded. In what follows, we assume that the actual values of the velocity $\actual{\mathbf{v}}(\cdot)$ are unknown, but guaranteed to lie in the known tube $[\mathbf{v}](\cdot)$. The loop detection problem then amounts to computing the set $\mathbb{T}$ containing all feasible loops according to the given uncertainties:
		\begin{equation}
			\mathbb{T}=\left\{ (t_1,t_2) \mid \exists \mathbf{v}(\cdot)\in[\mathbf{v}](\cdot), \int_{t_1}^{t_2} \mathbf{v}(\tau)d\tau=\mathbf{0} \right\},
		\end{equation}
		or equivalently:
		\begin{equation}
			\mathbb{T}=\left\{ (t_1,t_2) \mid \mathbf{0}\in [\mathbf{f}](t_1,t_2) \right\},
		\end{equation}
		with $[\mathbf{f}]:\mathbb{IR}^2\to\mathbb{IR}^2$ an inter-temporal inclusion function defined by
		\begin{equation}
			[\mathbf{f}]\left([t_1],[t_2]\right)=\int_{[t_1]}^{[t_2]}[\mathbf{v}](\tau)d\tau.
			\label{eq:loops:function:f}
		\end{equation}
		
		Hence, $\mathbb{T}$ is a reliable enclosure of $\actual{\mathbb{T}}$ so that for each $t$-pair in $\mathbb{T}$, there exist values in the set of measurements that lead to the detection of a feasible loop.
		Therefore, the following relation is guaranteed:
		\begin{equation}
			\actual{\mathbb{T}}\subseteq\mathbb{T}\subseteq[t_0,t_f]^2.
		\end{equation}
		
\begin{figure}[t!]
			\centering
			\subfloat[Loop detection over an undeniable looped trajectory.]
			{
				\includegraphics[width=0.8\linewidth]{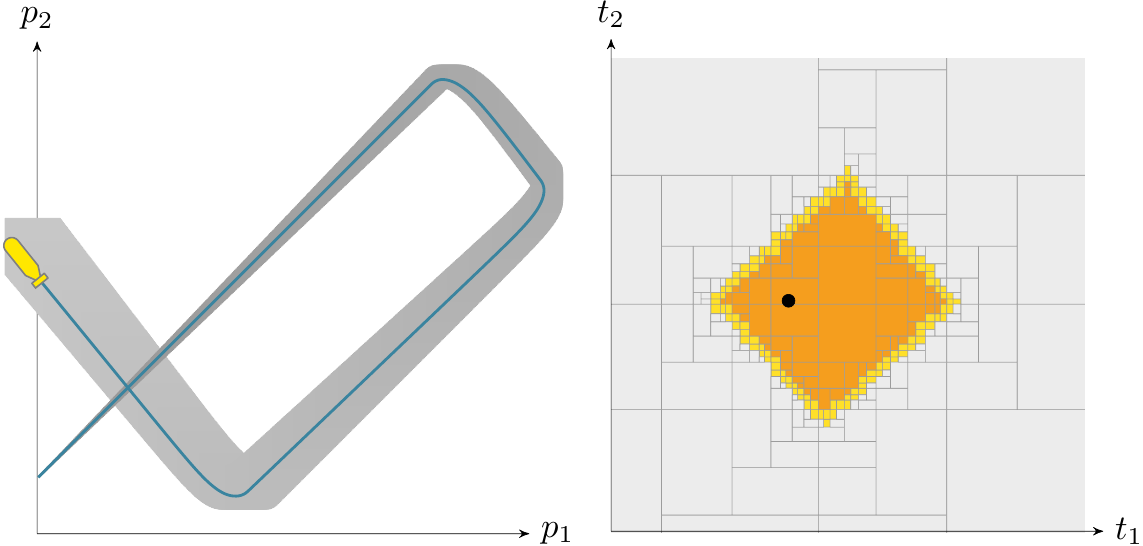}
				\label{fig:tpair:uncertainty:1}
			}
			
			\subfloat[Loop detection over a doubtful looped trajectory. In this case the actual trajectory is made of two loops approximated within the same detection.]
			{
				\includegraphics[width=0.8\linewidth]{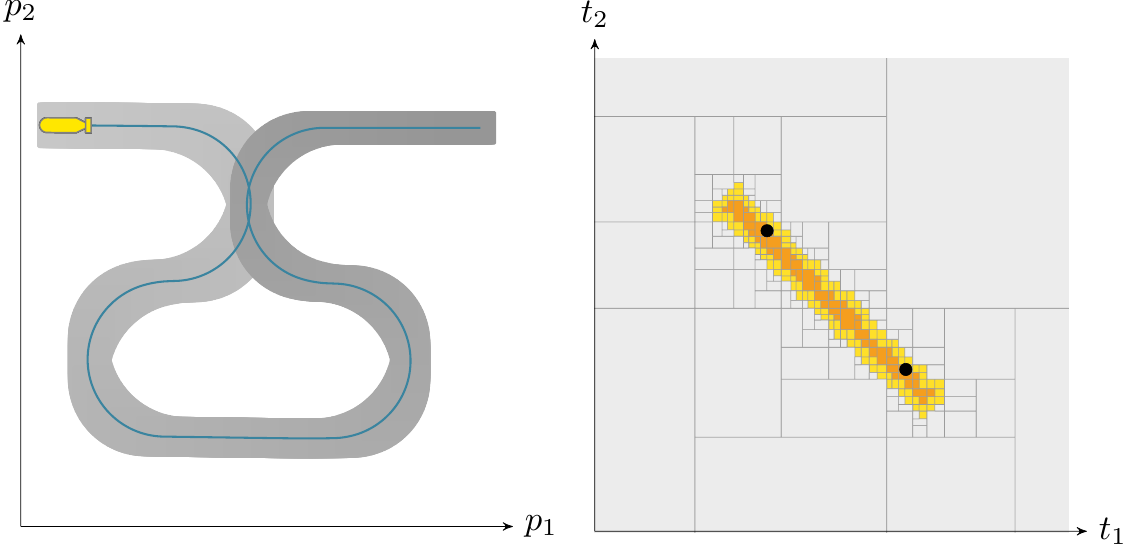}
				\label{fig:tpair:uncertainty:2}
			}
			
			\subfloat[Loop detection over a doubtful looped trajectory. In this case the actual trajectory never crosses itself despite a loop detection.]
			{
				\includegraphics[width=0.8\linewidth]{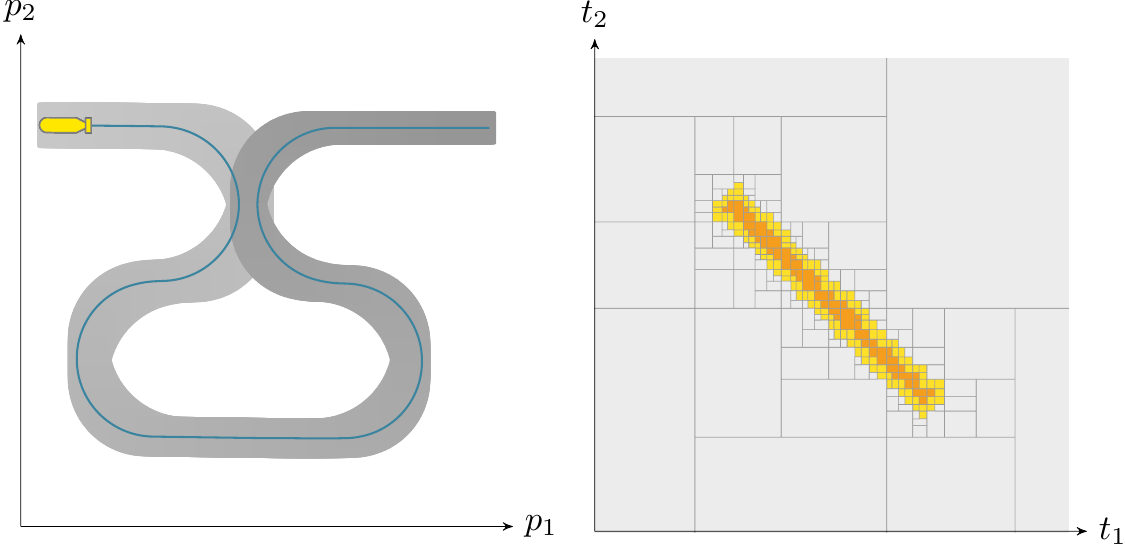}
				\label{fig:tpair:uncertainty:3}
			}
			\caption{Guaranteed loop detections of a mobile robot. Its evolution is drawn on the left hand side: the true trajectory is plotted in blue while the computed envelope of all feasible trajectories is represented in gray, thus depicting an increasing localization uncertainty due to strong measurement errors. A part of the $t$-plane is pictured on the right hand side with the loop detection set $\mathbb{T}$ approximated by a set of boxes $[\mathbf{t}]_i$. This reliable approximation is obtained with a SIVIA algorithm. When an actual loop $(t_1,t_2)$ exists -- pictured by a black dot -- it is surely enclosed by this set of boxes.}
			\label{fig:tpair:uncertainty}
		\end{figure}

		Figure~\ref{fig:tpair:uncertainty} illustrates numerical approximations of $\mathbb{T}$ with a SIVIA algorithm \cite{JaulinWalter93SetInvAutom,aubry:autom:13} over several examples. As can be seen, the detection of a potential loop is not a proof of its existence. For instance, Figures~\ref{fig:tpair:uncertainty:2}--\ref{fig:tpair:uncertainty:3} are two identical cases regarding the uncertainties: the detection $\mathbb{T}$ pictured in the $t$-plane is the same while the actual trajectory may let appear one loop, two loops, or none.

\begin{figure}[!h]
			\centering
			\includegraphics[width=1.0\linewidth]{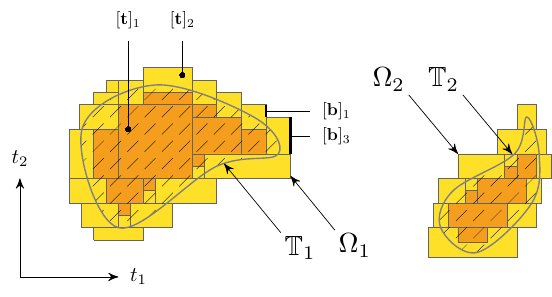}
			\caption{Approximation of a set $\mathbb{T}=\mathbb{T}_1\cup\mathbb{T}_2$ with sets of non-overlapping boxes. In this paper, only the outer approximations $\Omega_i$ (unions of connected boxes called \emph{subpavings}) will be assessed.}
			\label{fig:subpaving}
		\end{figure}

		Note that depending on robot's trajectory, the numerical approximation of $\mathbb{T}$ may consist of several connected components denoted $\mathbb{T}_i$, see Figure~\ref{fig:subpaving}. 
		
		The only way to prove the existence of at least one loop in a given subset $\mathbb{T}_i$ is to verify that $\forall\mathbf{f}\in[\mathbf{f}],\exists(t_1,t_2)\in\mathbb{T}_i$ such that $\mathbf{f}(t_1,t_2)=\mathbf{0}$, which is equivalent to verifying a zero of an unknown function $\actual{\mathbf{f}}\in[\mathbf{f}]$ on $\mathbb{T}_i$.
		This can be shown using the Newton test $\mathcal{N}$ from \cite{Moore79} or the new test $\mathcal{T}$ based on topological degree that will be presented in Sections \ref{sec:topodegree} and  \ref{sec:test:existence}.

\section{Topological degree for\\zeros verification}
\label{sec:topodegree}

In what follows, we assume that an inclusion function $[\mathbf{f}]: \mathbb{IR}^n\to\mathbb{IR}^n$ of the unknown continuous function $\actual{\mathbf{f}}: \R^n\to\R^n$ is given, possibly in the form of an algorithm for
computing $[\mathbf{f}]([\mathbf{t}])$.

We want to isolate and verify zeros of $\actual{\mathbf{f}}$.
It immediately follows from the definition that if $\mathbf{0}\notin [\mathbf{f}]([\mathbf{t}])$ for some box $[\mathbf{t}]$, then $\actual{\mathbf{f}}$ has no zero on $[\mathbf{t}]$.
It is, however, harder to verify the \emph{existence} of zero inside a region.
If $\mathbf{0}\in [\mathbf{f}]([\mathbf{t}])$, we cannot disprove $\actual{\mathbf{f}}(\mathbf{t})=\mathbf{0}$ for some $\mathbf{t}$, but it is also not obvious how to
prove the existence of such $\mathbf{t}$.

A~powerful tool for verifying zeros is the topological degree $\mathrm{deg}(\actual{\mathbf{f}},\Omega)$.
It is a~unique integer assigned to $\actual{\mathbf{f}}$ and a compact set\footnote{In some references such as~\cite{fonseca1995degree}$, 
\Omega$ is assumed to be open and bounded, which corresponds to considering the \emph{interior} of our $\Omega$. The requirement $\actual{\mathbf{f}}(\mathbf{t})\not=\mathbf{0},\forall\mathbf{t}\in\partial\Omega$ is unchanged.}
$\Omega\subset \R^n$ such that $\actual{\mathbf{f}}(\mathbf{t})\not=\mathbf{0}$ for all $\mathbf{t}\in \partial\Omega$.
%
%
%
%
The topological degree satisfies certain properties, see~\cite{fonseca1995degree,cho2006topological,Furi2010} for detailed expositions. 
For our purposes, the most important property is that
\begin{equation}
\label{e:degree2zero}
\deg(\actual{\mathbf{f}},\Omega)\neq 0 \quad\implies \quad\exists \mathbf{t}\in \Omega~\mid~\actual{\mathbf{f}}(\mathbf{t})=\mathbf{0}.
\end{equation}

\begin{figure}[!h]
	\centering
	\includegraphics[width=\linewidth]{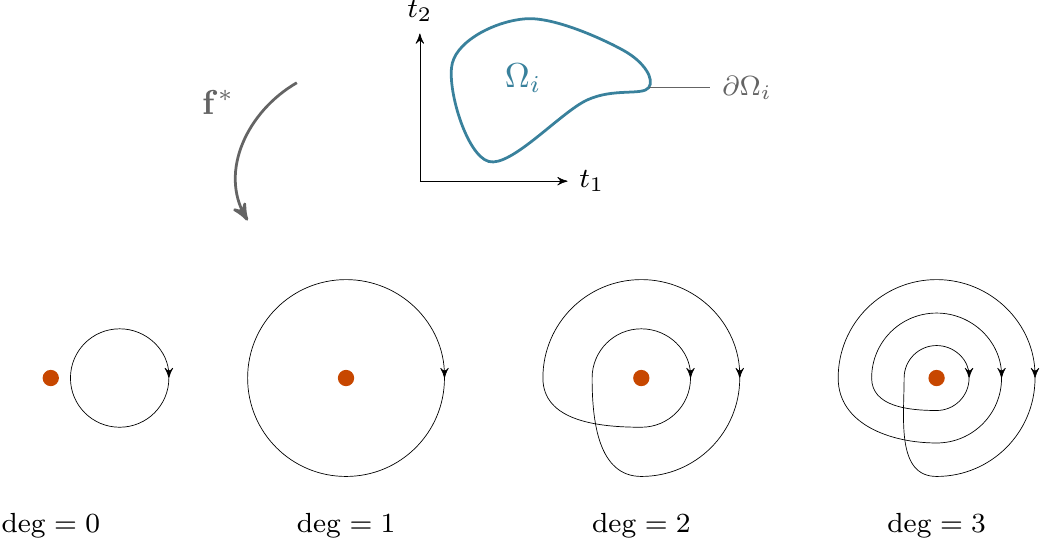}
	\caption{Computation of the degree of $\actual{\mathbf{f}}$ on $\Omega_i$. The illustration shows several positive degree cases.}
	\label{fig:degree:winding}
\end{figure}

Recent advances in computational topology generated many algorithms for computing the topological degree. Besides, it can be computed
in case where only an inclusion function $[\mathbf{f}]$ of $\actual{\mathbf{f}}$ is given. It was argued in~\cite[Sec. 9]{quasi} that the degree test is in many cases more
powerful than more classical verification tools including interval Newton, Miranda's or Borsuk's tests (see~\cite{moore1977test,moore1980simple,borsuk1933drei} for definitions of those tests).
%
Our application for detecting robot loops deals with the case $n=2$.
Then the degree has a particularly nice geometric
interpretation: it is the \emph{winding number} of the curve $\partial \Omega\stackrel{\actual{\mathbf{f}}}{\mapsto} \R^2\setminus\{\mathbf{0}\}$ around $\mathbf{0}$, see Figure~\ref{fig:degree:winding}.
If $[\mathbf{f}]$ is given, then the winding number can be computed by a number of elementary methods, the algorithm of~\cite{franek2012} being one of them.

Consider a given subdomain $\mathbb{T}\subset \R^n$ in which we want to find zeros of $\actual{\mathbf{f}}$. For computational purposes, an outer approximation of $\mathbb{T}$ is performed by dividing the space into a set of non-overlapping boxes denoted $[\mathbf{t}]_j$.
An algorithm relying on set inversion such as SIVIA \cite{JaulinWalter93SetInvAutom} can be used to this end. Figure~\ref{fig:subpaving} depicts such reliable approximation.
The outer set has the properties required for $\Omega$. Consequently, the set $\Omega$ we consider will always be a~finite union of boxes.

The following statement is a~reformulation of~\cite[Theorem 2.9]{franek2012} adapted to our notation.
\begin{theorem} 
	\label{th:topDeg:interval}
	Let $\Omega$ be a~union of finitely many boxes in $\mathbb{IR}^n$:
	\begin{equation}
		\Omega=\bigcup_{j=1}^l [\mathbf{t}]_j,
	\end{equation}
	and assume that the boundary $\partial\Omega$ is a~union of finitely many boxes
	\begin{equation}
		\partial\Omega=\bigcup_{k=1}^p [\mathbf{b}]_k~\footnote{We also consider degenerate boxes. In this case, $[\mathbf{b}]$'s are boxes in $\mathbb{IR}^n$ of topological dimension $n-1$.}.
	\end{equation}
	If  $\mathbf{0}\notin [\mathbf{f}]([\mathbf{b}]_k)$ for all $k=1,\ldots, p$, then the degree $\mathrm{deg}(\actual{\mathbf{f}},\Omega)$ is uniquely determined and computable only from 
	the evaluations $[\mathbf{f}]([\mathbf{b}]_k)$.
\end{theorem}
It immediately follows that, under the assumptions of the Theorem, $\mathrm{deg}(\mathbf{g},\Omega)=\mathrm{deg}(\actual{\mathbf{f}},\Omega)$ for any $\mathbf{g}\in [\mathbf{f}]$, because $[\mathbf{f}]$ is also an inclusion function for $\mathbf{g}$ in such case.

Let $\Omega_1,\ldots,\Omega_l$ be connected components of the union of such boxes $[\mathbf{t}]$ with potential zeros. On each $\Omega_i$, if its boundary is covered by
boxes $[\mathbf{b}]_k$ such that $\mathbf{0}\notin [\mathbf{f}]([\mathbf{b}]_k)$ for each $k$, we can compute the degree $\mathrm{deg}(\actual{\mathbf{f}},\Omega_i)$.
Whenever this is nonzero, we verified the existence of at least one $\mathbf{t}\in \Omega_i$ such that $\actual{\mathbf{f}}(\mathbf{t})=\mathbf{0}$. 
We emphasize that the function $\actual{\mathbf{f}}$ was unknown and we only worked with its inclusion function $[\mathbf{f}]$.

In the above paragraph, we never used derivatives of $\mathbf{f}^*$.
Using additional information on derivatives, we can also count the number of solutions.
Namely, if $\Omega$ is connected and $\mathrm{deg}(\actual{\mathbf{f}},\Omega)=\ell$
and we further know that the Jacobian matrix $\mathbf{J}_{\actual{\mathbf{f}}}$ is nonsingular everywhere on $\Omega$, then $\actual{\mathbf{f}}$ has \emph{exactly} $|\ell|$
solutions in $\Omega$. This immediately follows from the definition of the degree given, for example, in~\cite[p. 27]{Milnor}.
In particular, if the degree is $\pm 1$, then non-singularity immediately implies that there is a unique zero of $\actual{\mathbf{f}}$ in $\Omega$.
More details about the implementation of this is given in Section~\ref{sec:test:uniqueness}.

\section{Loop existence test}
\label{sec:test:existence}

The topological degree theory will be used for proving the existence of robot loops. This section provides the proposed \emph{existence test} with 
an explicit algorithm.

\subsection{From topological degree\\to loops proofs}

The inclusion function $[\mathbf{f}]$ assumed in Section \ref{sec:topodegree} is given by Eq.~\eqref{eq:loops:function:f}, while its 
computation is based on Eq.~\eqref{eq:tubes:boundedbounds}. A SIVIA algorithm relying on Eq. \eqref{eq:loops:function:f} provides an outer approximation $\Omega$ of the set $\mathbb{T}$ resulting in several subpavings denoted by $\Omega_i$. Such algorithm provides guaranteed results given the inclusion function that can be built from datasets, see \cite{JaulinWalter93SetInvAutom}. The following relation is then guaranteed:
\begin{equation}
\actual{\mathbb{T}}\subset\mathbb{T}\subset\left(\bigcup_{i} \Omega_i\right)\subset[t_0,t_f]^2.
\end{equation}

Each of these subpavings $\Omega_i$ constitutes a potential loop detection: 
there exists at least one trajectory with a $\mathbf{v}(\cdot)\in [\mathbf{v}](\cdot)$ that looped for one $t$-pair belonging to $\Omega_i$. 
However, the trajectory related to the actual but unknown $\actual{\mathbf{v}}(\cdot)$ may have never looped in reality despite the detection, 
as pictured by Figure~\ref{fig:tpair:uncertainty}.
As a consequence, proving a loop amounts to verifying a zero of $\actual{\mathbf{f}}:\mathbf{t}\mapsto\int_{t_1}^{t_2}\actual{\mathbf{v}}(\tau)d\tau$ 
in $\Omega_i$ using the known inclusion function given by Eq. \eqref{eq:loops:function:f}.
By using the topological degree in this context, the consequent of the implication given in Eq. \eqref{e:degree2zero} is a proof of a loop existence.
The algorithm for numerical verification of $\mathrm{deg}(\actual{\mathbf{f}},\Omega_i)\not=0$ is provided hereinafter.


\subsection{Implementation}

This section shows how to apply a~simple version of the topological degree algorithm for the special case of
a~connected two-dimensional region $\Omega_i$ that consists of 2D boxes. The following algorithms are an adaptation of~\cite{franek2012}
for this special case.

Assume that $\Omega_i\subset\R^2$ is a~union of finitely many boxes and the boundary $\partial\Omega_i$ is a topological circle\footnote{Hence, we shall assume the set $\Omega_i$ is strictly included in $[t_0,t_f]^2$ so that a closed boundary $\partial\Omega_i$ can be assessed.}.
Further, let $\mathbf{a}_1\ldots, \mathbf{a}_p$ be points in $\partial\Omega_i$ and $[\mathbf{b}]_1,\ldots, [\mathbf{b}]_p$
be edges covering the boundary $\partial\Omega_i$, such that $\partial [\mathbf{b}]_i=\{\mathbf{a}_{i+1}, \mathbf{a}_i\}$ for $i<p$ and
$\partial [\mathbf{b}_p]=\{\mathbf{a}_1,\mathbf{a}_p\}$. We endow each $[\mathbf{b}]_i$ with an \emph{orientation} such that
$\mathbf{a}_{i+1}$ is an end-point of $[\mathbf{b}]_i$ and $\mathbf{a}_i$ is the starting-point of $[\mathbf{b}]_i$ for $i<p$ and,
similarly, $\mathbf{a}_1$ is the end-point of $[\mathbf{b}]_p$ and $\mathbf{a}_p$ the starting-point of $[\mathbf{b}]_p$.
We define the \emph{oriented boundary} of $[\mathbf{b}]_i$ to be $\mathbf{a}_{i+1}-\mathbf{a}_i$ for $i<p$ and the oriented boundary of $[\mathbf{b}]_p$
to be $\mathbf{a}_1-\mathbf{a}_p$, where we introduce \emph{oriented vertices} $\pm \mathbf{a}_j$ as formal symbols. This structure of oriented edges
and oriented vertices can easily be represented in a computer.

%
%

Further, assume that an interval function $[\mathbf{f}]$ is given such that $\mathbf{0}\notin [\mathbf{f}]([\mathbf{b}]_i)$ for all $i$.
This means that either the first or the second coordinate of this box has a constant sign, $+$ or $-$.
We assign to the oriented box $[\mathbf{b}]_i$ the pair $(c_i, s_i)$ where $c_i\in \{1,2\}$ and $s_i\in \{+,-\}$
in such a~way that the $c_i$-th coordinate of $[\mathbf{f}]([\mathbf{b}]_i)$ has a constant sign $s_i$.
For example, $(2,-)$ indicates that the second coordinate of $[\mathbf{f}]([\mathbf{b}]_i)$ is negative: in particular $\actual{f_2}$
is negative on $[\mathbf{b}]_i$.
Such choice $(c_i, s_i)$ is not necessarily unique, but any choice will give us a~correct result at the end.

The degree $\mathrm{deg}(\actual{\mathbf{f}},\Omega_i)$ can be computed using the following algorithms. The existence test $\mathcal{T}$ is then a direct conclusion on the computed degree. One should note that, at this step, the Algorithm \ref{alg:existence:test} is not able to reject the feasibility of a loop. In case of a non-zero degree, it will prove a loop existence. Otherwise, the $``\varnothing"$ output will reflect a non-conclusive test.

\begin{algorithm}[h]
	\begin{algorithmic}[1]
		\State \textbf{begin}
		\State $[\mathbf{b}]_1\dots[\mathbf{b}]_p\leftarrow\mathrm{getContour}\left(\Omega_i\right)$
		\If{$\mathrm{2dTopoDegree}\left([\mathbf{b}]_1\dots[\mathbf{b}]_p,[\mathbf{f}]\right)\not=0$}
			\Return $\mathrm{true}$
		\Else
			\Return $\varnothing$ \quad \emph{// not able to conclude about existence}
		\EndIf
		\State \textbf{end}
	\end{algorithmic}
	
	\caption{$\mathrm{existenceTest}\mathcal{T}\left(\mathrm{in}:\Omega_i,[\mathbf{f}] - \mathrm{out}:\mathrm{true}|\varnothing\right)$}
	\label{alg:existence:test}
\end{algorithm}

\begin{algorithm}[h]
	\begin{algorithmic}[1]
		\State \textbf{begin}
		\State $d\leftarrow 0$
		\For{$i=1$ to $p$}
			\State $(c_i,s_i)\leftarrow\mathrm{tagEdge}\left([\mathbf{b}]_i,[\mathbf{f}]\right)$
		\EndFor
		\State $c_0\leftarrow c_p$, $s_0\leftarrow s_p$, $c_{p+1}\leftarrow c_1$, $s_{p+1}\leftarrow s_1$
		\For{$i=1$ to $p$}
			\If{$(c_i,s_i)=(1,+)$}
				\If{$(c_{i+1}, s_{i+1})=(2,+)$}
					\State $d\leftarrow d+1$
				\EndIf
				\If{$(c_{i-1}, s_{i-1})=(2,+)$}
					\State $d\leftarrow d-1$
				\EndIf
			\EndIf
		\EndFor
		\Return $d$
		\State \textbf{end}
	\end{algorithmic}
	
	\caption{$\mathrm{2dTopoDegree}\left(\mathrm{in}:[\mathbf{b}]_1\dots[\mathbf{b}]_p,[\mathbf{f}] - \mathrm{out}:d\right)$}
	\label{alg:topdeg2d}
\end{algorithm}

\begin{algorithm}[h]
	\begin{algorithmic}[1]
		\State \textbf{begin}
		\If{$0\not\in[f_1]([\mathbf{b}])$}
			\State \textbf{if} $[f_1]([\mathbf{b}])\subset\mathbb{R}^{+}$, \textbf{return} $(1,+)$
			\State \textbf{else}, \textbf{return} $(1,-)$
		\ElsIf{$0\not\in[f_2]([\mathbf{b}])$}
			\State \textbf{if} $[f_2]([\mathbf{b}])\subset\mathbb{R}^{+}$, \textbf{return} $(2,+)$
			\State \textbf{else}, \textbf{return} $(2,-)$
		\Else
			\Return $\varnothing$ \quad \emph{// note: this case should not happen}
		\EndIf
		\State \textbf{end}
	\end{algorithmic}
	
	\caption{$\mathrm{tagEdge}\left(\mathrm{in}:[\mathbf{b}],[\mathbf{f}] - \mathrm{out}: (c,s)\right)$}
	\label{alg:tag:edge}
\end{algorithm}

\begin{figure}[!h]
	\begin{center}
		\includegraphics[width=0.6\linewidth]{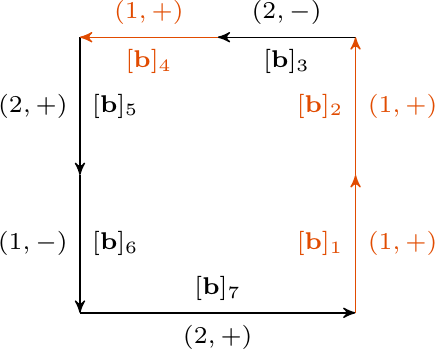}
	\end{center}
	\caption{Illustration of the degree algorithm. The \emph{selected} edges in this case are $[\mathbf{b}]_1, [\mathbf{b}]_2, [\mathbf{b}]_4$ but only $[\mathbf{b}]_1$ results in an addition by $-1$ and $[\mathbf{b}]_4$ in an addition of $+1$. The overall degree is zero in this case.}
	\label{fig:degree_comp}
\end{figure}

An illustration of Algorithm \ref{alg:topdeg2d} is given in Figure~\ref{fig:degree_comp}. Here the algorithm returns zero, because the if-conditions are satisfied only for
the edge $[\mathbf{b}]_1$ where $d$ will change from $0$ to $-1$, and then in edge $[\mathbf{b}]_4$ where $d$ will be changed from $-1$~to~$0$.

If our representation of $\Omega_i$ comes from the previous SIVIA algorithm, we can assume that the \verb|getContour| function 
(in Algorithm~\ref{alg:existence:test}) is available and has linear time-complexity. 
A naive implementation of Algorithm~\ref{alg:topdeg2d} has quadratic complexity. Its input $[\mathbf{b}]_1,\ldots [\mathbf{b}]_p$ 
can be ordered and oriented in $\sim p^2$ steps so that the end-point of $[\mathbf{b}]_j$ (resp. $[\mathbf{b}]_p$) coincides with the starting-point 
of $[\mathbf{b}]_{j+1}$ (resp. $[\mathbf{b}]_1$). The rest then amounts to finding the signs $(c_j,s_j)$ in one pass over all $j$ and adding $1$ (resp. $-1$) to a~global variable whenever $(c_j, s_j)=(1,+)$ and the next (resp. previous) sign is  $(2,+)$.
A~better implementation in $O(p)$ is possible if we can access additional information, such as
the boundary orientation of $[\mathbf{b}]_j$ induced from $\partial\Omega_i$.

\section{Reliable number of loops}
\label{sec:test:uniqueness}

Aside from proving the existence of a loop, it may be interesting to count the number of solutions. This can be done using additional information on the derivatives. To this end, the Jacobian matrix $\mathbf{J}_{\actual{\mathbf{f}}}$ of the unknown $\actual{\mathbf{f}}$ has to be approximated by $\left[\mathbf{J}_{{\mathbf{f}}}\right]$. From Leibniz integral rule,
\begin{equation}
\left[\mathbf{J}_{{\mathbf{f}}}\right]\left([\mathbf{t}]\right)=
\begin{pmatrix}
	\frac{\partial [f_1]}{\partial [t_1]} & \frac{\partial [f_1]}{\partial [t_2]} \\[1.5ex]
	\frac{\partial [f_2]}{\partial [t_1]} & \frac{\partial [f_2]}{\partial [t_2]}
\end{pmatrix}
=
\begin{pmatrix}
-[v_1]([t_1]) & [v_1]([t_2]) \\[1.5ex]
-[v_2]([t_1]) & [v_2]([t_2])
\end{pmatrix},
\label{eq:jacobi:approx}
\end{equation}
where $[\mathbf{v}](\cdot)$ is the tube containing the unknown velocity $\actual{\mathbf{v}}(\cdot)$ of the robot.

If $\Omega_i$ is a compact set as defined in Section \ref{sec:topodegree} and if the Jacobian matrix $\mathbf{J}_{\actual{\mathbf{f}}}$ is nonsingular everywhere on $\Omega_i$, then the absolute value of the degree is the exact number of solutions for $\actual{\mathbf{f}}=\mathbf{0}$ in $\Omega_i$.

Proving the non-singularity of the Jacobian matrix amounts to verifying that its determinant is non-zero. Using the inclusion function from Eq. \eqref{eq:jacobi:approx}, this is equivalent to verifying $0\not\in\det\left(\left[\mathbf{J}_{{\mathbf{f}}}\right]\right)$.

The algorithm \ref{alg:loopsnumber:test} provided hereinafter returns the exact number of loops in a set $\Omega_i$ when the zeros are robust enough. Otherwise, nothing can be concluded regarding the uncertainties of the information.

\begin{algorithm}[h]
	\begin{algorithmic}[1]
		\State \textbf{begin}
		\State $[\mathbf{t}]_1\dots[\mathbf{t}]_j\leftarrow\mathrm{getBoxes}\left(\Omega_i\right)$
		\For{$k=1$ to $j$}
			\If{$0\in\det\left(\left[\mathbf{J}_{{\mathbf{f}}}\right]([\mathbf{t}]_k)\right)$}
				\Return $\varnothing$
			\EndIf
		\EndFor
		\State $[\mathbf{b}]_1\dots[\mathbf{b}]_p\leftarrow\mathrm{getContour}\left(\Omega_i\right)$
		\State $\ell\leftarrow \mathrm{2dTopoDegree}\left([\mathbf{b}]_1\dots[\mathbf{b}]_p,[\mathbf{f}]\right)$
		\Return $\left|\ell\right|$
		\State \textbf{end}
	\end{algorithmic}

\caption{$\mathrm{loopsNumber}\left(\mathrm{in}:\Omega_i,[\mathbf{f}],\left[\mathbf{J}_{{\mathbf{f}}}\right] - \mathrm{out}:\ell\right)$}
\label{alg:loopsnumber:test}
\end{algorithm}


\begin{remark}
	The algorithm used to compute the set $\Omega_i$ may provide wide boxes $[\mathbf{t}]_k$ that will result in an over-approximation of the $\left[\mathbf{J}_{{\mathbf{f}}}\right]([\mathbf{t}]_k)$. A bisection of the $[\mathbf{t}]_k$ may be applied when $0\in\det\left(\left[\mathbf{J}_{{\mathbf{f}}}\right]([\mathbf{t}]_k)\right)$ in order to deal with smaller boxes, thus reducing the pessimism of the Jacobian evaluation 
and increasing the chances to disprove $0\in\det\left(\left[\mathbf{J}_{{\mathbf{f}}}\right]([\mathbf{t}]_k)\right)$.
If the determinant approximation still contains $0$ beyond a given precision, then the algorithm should stop being not able to conclude.
\end{remark}

\section{Application on real datasets}
\label{sec:application}

	The efficiency of the proposed test is demonstrated over two experiments involving actual underwater robots. The underwater case is challenging as robots do not benefit from GPS fixes except at the very beginning of the mission. Hence, dead-reckoning methods usually apply for state estimation, leading to strong cumulative errors. Loops will be proven in this context.
	
	\subsection{Absolute velocities}
	
		Underwater robots are usually equipped with an Inertial Measurement Unit (IMU) providing the Euler angles $(\psi,\theta,\phi)$ depicting the orientation of the robot. In addition, a Doppler Velocity Log (DVL) will track the vehicle's speed $\mathbf{v}_r\in\mathbb{R}^3$ over the seabed by acoustic means, providing values in robot's own coordinate system. The absolute speed vector $\mathbf{v}\in\mathbb{R}^3$, expressed in the environment reference frame, is then obtained by
		\begin{equation}
			\mathbf{v} = \mathbf{R}(\psi,\theta,\varphi)\ \cdot\ \mathbf{v}_{r},
		\end{equation}
		where $\mathbf{R}(\psi,\theta,\varphi)$ is a classical Euler matrix. For more details about state equations for underwater robots, one can refer to \cite{fossen:guidance,JaulinIEEETRO09}.

	\subsection{From sensors to reliable results}
	
		
		\subsubsection{Obtaining bounded measurements at time $t$\\}
			
			In practice, a measurement error is often modeled by a Gaussian distribution which has an infinite support. Therefore, setting bounds around this measurement already constitutes a theoretical risk of loosing the actual value. A choice has to be made at this step, considering such risk. After that, however, any algorithm standing on interval methods is ensured to not increase this risk.
			
			Data-sheets usually give sensor specifications such as the standard deviation $\sigma$. Hence, a measurement $v_1$ is assumed to belong to an interval $[v_1]$ centered on $v_1$ and inflated according to the sensor uncertainties. For instance, $[v_1]=[v_1-2\sigma,v_1+2\sigma]$ will provide a $95\%$ confidence rate over the actual and unknown value $\actual{v}_1$, considering the Gaussian distribution.
			
%

		\subsubsection{From measurements to tubes\\}
		\label{sub:sensor2tube}
		
			
			
			Common sensors provide us only with a~set of measurement vectors sampled over finitely many time values, while our algorithm deals with continuous interval functions. Our choice is to build a tube from this data by computing a~piecewise linear interpolation $\mathbf{v}^{PL}(t)$ between the measurements. We then create a tube $[\mathbf{v}](\cdot)$ such that\footnote{In fact, in our implementation, we enclose $\mathbf{v}^{PL}(\cdot)$ by an~even larger neighborhood. Our choice is to build the tube as a set of boxes representing slices. We first subdivide $[t_0, t_f]$ into a set of small sub-intervals $[t_k,t_{k+1}]$ corresponding to groups of several velocity measurements. We then define each slice as a box $[t_k,t_{k+1}]\times\left([-2\sigma, 2\sigma]^2+\cup_{t=t_k}^{t_{k+1}}\mathbf{v}^{PL}(t)\right)$. }
			\begin{equation}
				[\mathbf{v}](\cdot)=\mathbf{v}^{PL}(\cdot)+[-2\sigma,2\sigma]^2.
			\end{equation}
			Note that some sensors may provide real-time evaluations of $\sigma$, depending on the uncertainties of the environment\footnote{With DVL for instance, the velocity estimations are highly related to the altitude of the sensor over the seabed and the assumed knowledge of the water column, through which acoustic signals are propagated.}. In this case, $[\mathbf{v}](\cdot)$ can also be built with a reliable non-constant thickness.
			
\begin{figure}[!h]
	\centering
	\includegraphics[width=1.0\linewidth]{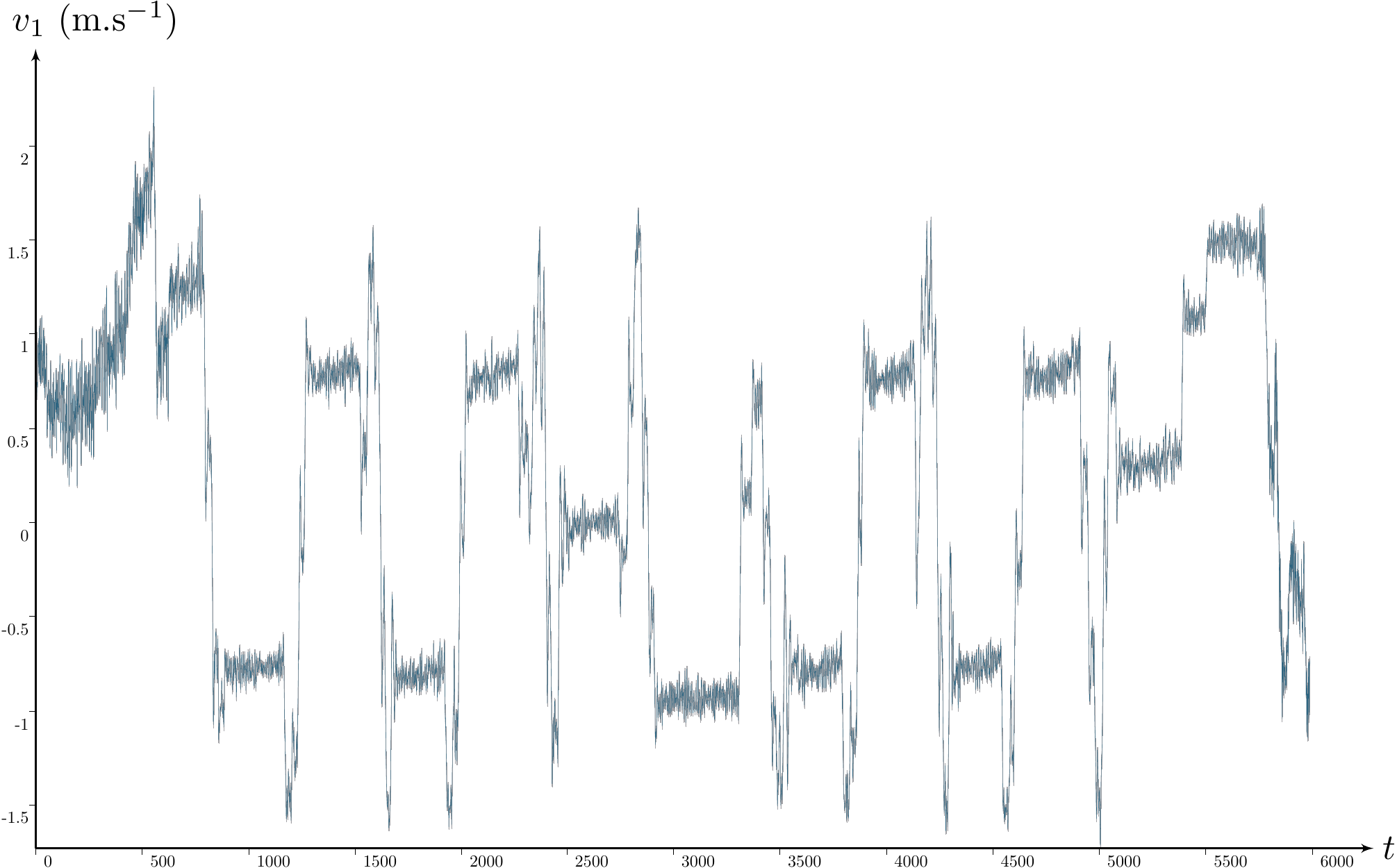}
	\caption{Tube $[v_1](\cdot)$ enclosing the \emph{Redermor} East velocity. Despite appearances, the signal is not noised: the temporal domain of 1H40 is compressed to fit the tube in the figure.}
	\label{fig:redermor:east:velocity}
\end{figure}

Practically, the time-sampling is much finer than any sudden velocity change and it is realistic to assume that the error 
$\mathbf{v}^{PL}(t)- \actual{\mathbf{v}}(t)$ is approximately normally distributed and centered at zero. An example of a tube $[v_1](\cdot)$ is provided in Figure~\ref{fig:redermor:east:velocity}.

The interval function $[\mathbf{f}]$ used for loop detection is then defined \textit{via} Equation~\eqref{eq:loops:function:f} as the integral of $[\mathbf{v}]$.

Our method for loop detection is reliable under the assumption $\mathbf{f^*}([\mathbf{t}])\subseteq [\mathbf{f}]([\mathbf{t}])$.
This inclusion immediately follows from the assumption $\mathbf{v^*}(\cdot) \subseteq [\mathbf{v}](\cdot)$ but in fact, 
the former inclusion is much more robust with respect to random velocity errors than the latter.\footnote{The 
real displacement $\int_{t_a}^{t_b} \actual{\mathbf{v}}(\tau)\,d\tau$ 
could lie outside $[\mathbf{f}](t_a,t_b)$ only if the velocity errors would \emph{cumulate in one direction}. 
More precisely, the projection of $\mathbf{v}^{PL}(\cdot)-\actual{\mathbf{v}}(\cdot)$ into one particular 
direction would have to be at least $2\sigma$ \emph{in average}, over the whole time interval $[t_a, t_b]$.
Under fairly general assumptions on the distribution of the velocity errors, such probability decreases exponentially with $(t_b-t_a)$.}
A~quantitative analysis of error probabilities is a~work in progress.

	
	\subsection{The \emph{Redermor} mission}

\begin{figure}[!h]
			\centering
			\includegraphics[width=0.9\linewidth]{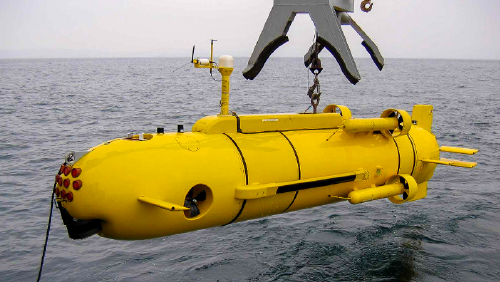}
			\caption{The \emph{Redermor} autonomous underwater robot before a sea trial. This experiment has been done with the kind help of \emph{DGA Techniques Navales Brest} (French Ministry of Defense).}
			\label{fig:redermor}
		\end{figure}

		This first application involves an Autonomous Underwater Vehicle (AUV) named \emph{Redermor}, see Figure~\ref{fig:redermor}. This test case has already been the subject of \cite[Sec. 6]{aubry:autom:13}, in which the existence of $14$ loops had been proved thanks to the test $\mathcal{N}$ relying on the Newton operator. Our goal is to compare these results with the topological degree test $\mathcal{T}$ we propose in this paper.
		
\begin{figure}[!h]
			\includegraphics[width=1.0\linewidth]{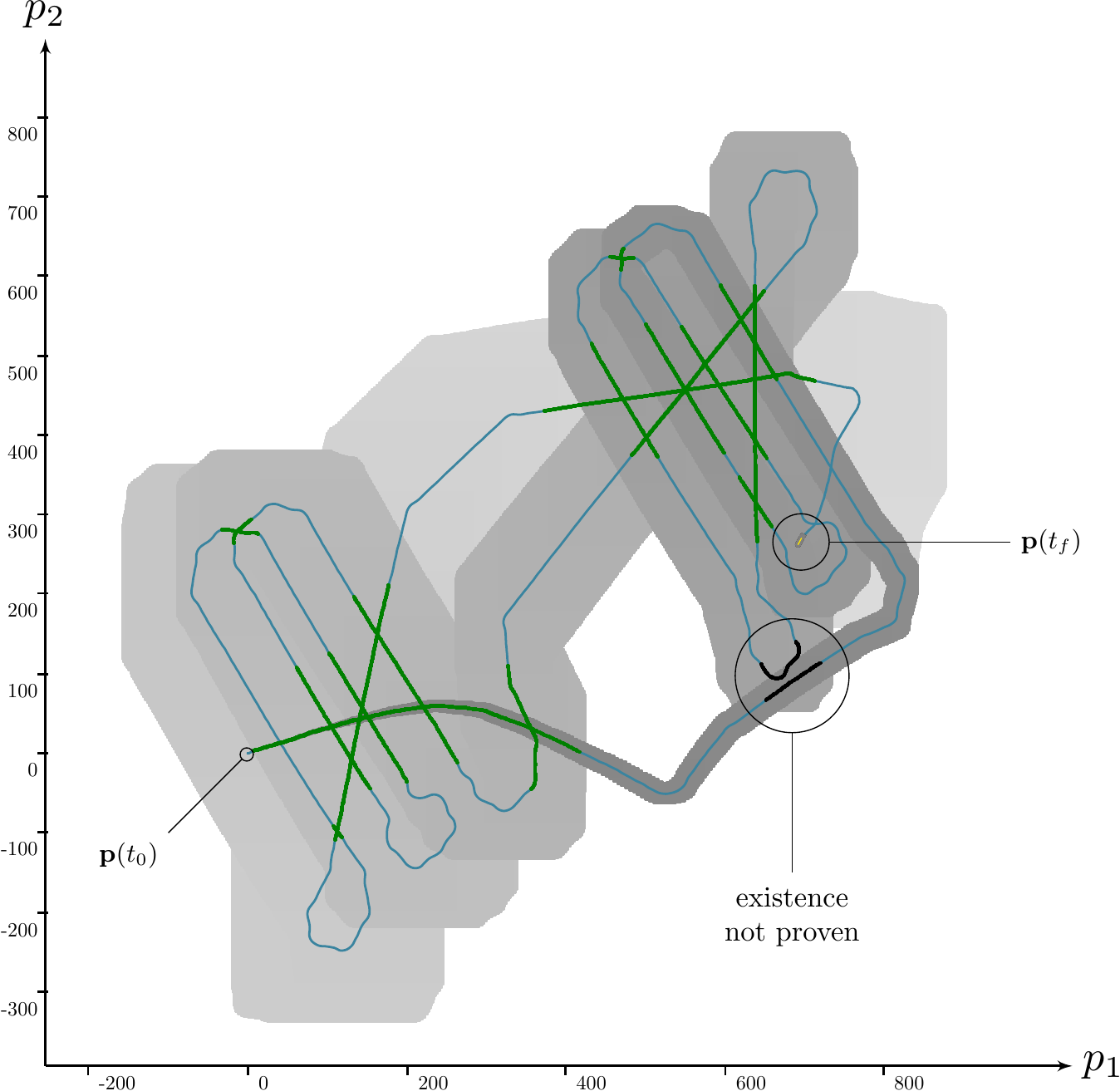}
			\caption{2D trace of \emph{Redermor} AUV. The projected tube $[\mathbf{p}](\cdot)$ (\emph{i.e.} the bounded estimated positions) is drawn in gray, depicting an increasing localization uncertainty. The truth is plotted by the blue line while green and black lines are the projections of the results given by the topological degree test $\mathcal{T}$.}
			\label{fig:redermor:positions}
		\end{figure}

\begin{figure}[!h]
			\includegraphics[width=1.0\linewidth]{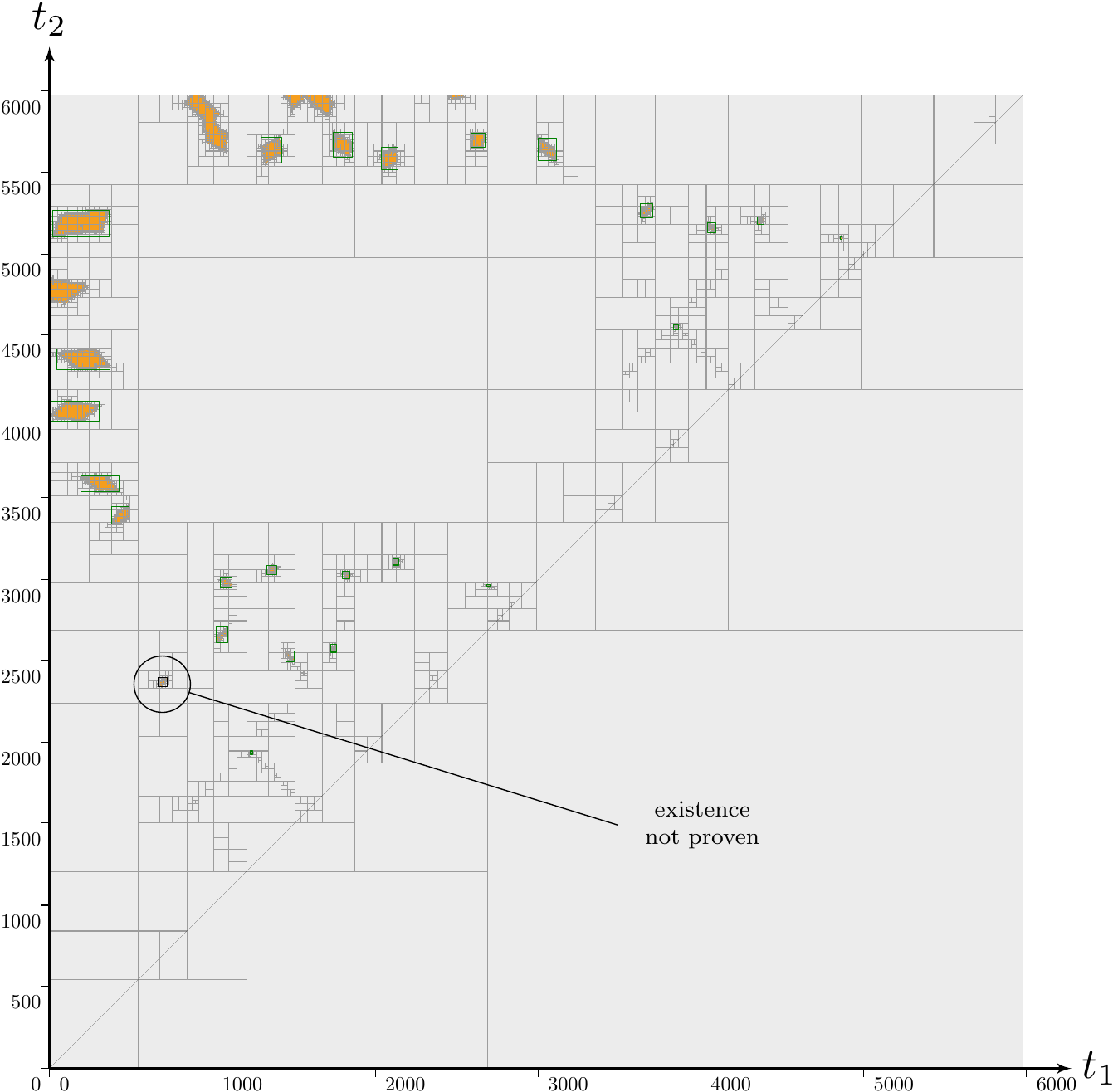}
			\caption{$t$-plane corresponding to the \emph{Redermor} mission and computed with a SIVIA algorithm. There exist four partial detections $\Omega_i$ on $t$-plane's edges that will not be considered here since the $\partial\Omega_i$ are not totally defined. They represent possible loops performed at the very beginning of the mission ($t_1\simeq t_0$) or at the end ($t_2\simeq t_f$). The diagonal line corresponds to the \emph{no-delay} line for which $t_1=t_2$.}
			\label{fig:redermor:tplane}
		\end{figure}

		A two hours experimental mission has been done in the Douarnenez bay in Brittany (France). A top view of the area covered by the robot is pictured in Figure~\ref{fig:redermor:positions}. \emph{Redermor} performed 28 loops, $20$m deep. The set-membership approach provides the enclosure of $\actual{\mathbf{v}}(\cdot)$, see Figure~\ref{fig:redermor:east:velocity}, and then the approximation of $\mathbb{T}$ pictured in the $t$-plane of Figure~\ref{fig:redermor:tplane}. A total of $25$ complete loop-detection sets have been computed on this test-case, the other solutions being partial. By \emph{complete detections} we mean loop detection sets $\Omega_i$ strictly included in the $t$-plane. Further comments on this application will only stand on these detections and the related actual loops.
		
\begin{figure}[!h]
			\centering
			\includegraphics[width=1.0\linewidth]{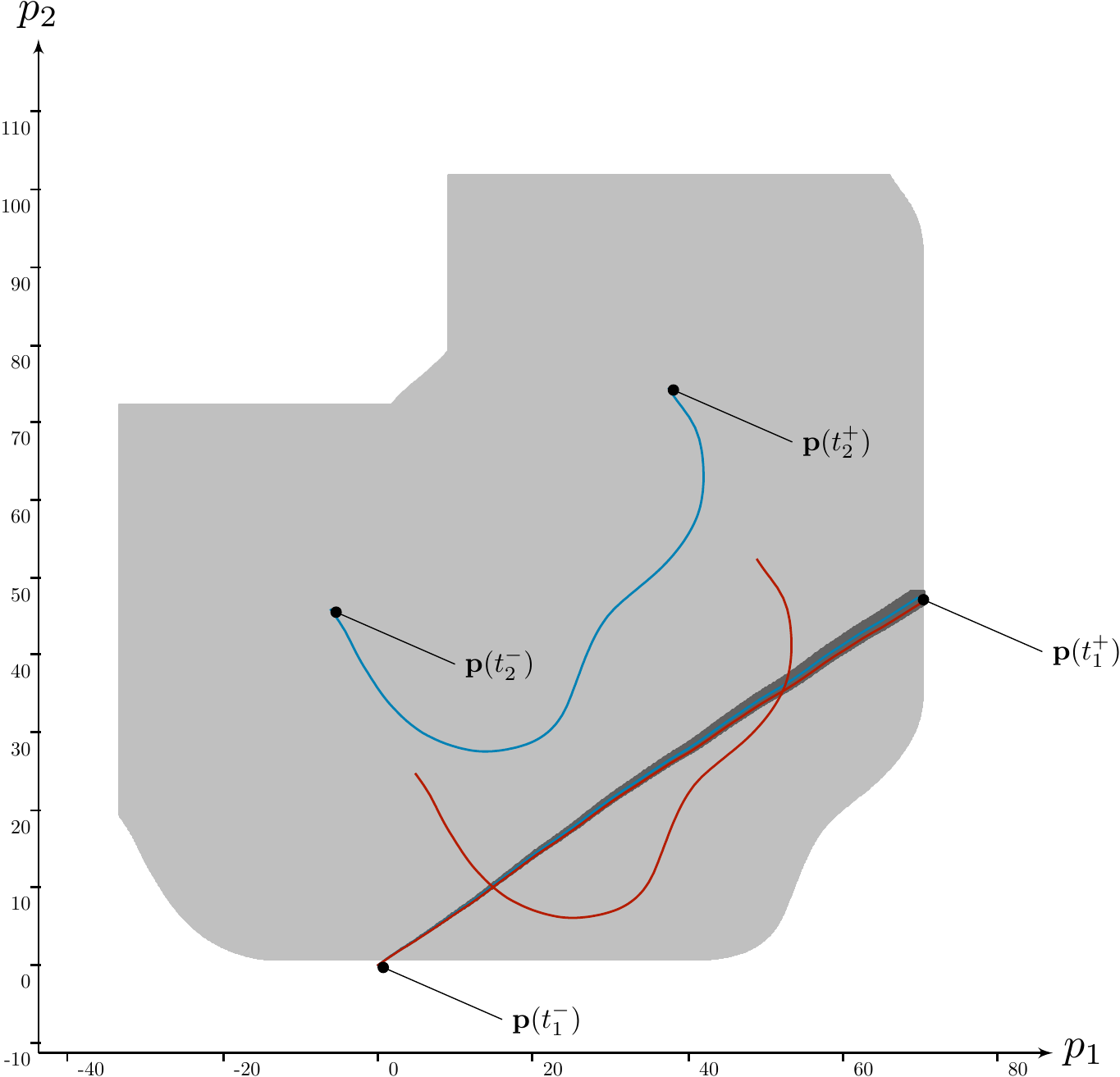}
			\caption{Independent projection of the non-conclusive case. Let us consider the loop-box $[\lb{t_1},\ub{t_1}]\times[\lb{t_2},\ub{t_2}]$ enclosing the corresponding $\Omega_i$ approximation. The actual trajectory over both $[\lb{t_1},\ub{t_1}]$ and $[\lb{t_2},\ub{t_2}]$ is plotted in blue. The bounded approximation of it is pictured in dark gray for the first part and light gray then. Note that we do not represent the amount of uncertainties gathered before $\lb{t_1}$: $\mathbf{p}(\lb{t_1})$ is well-known in this independent view. However, the amount of uncertainties over $[\lb{t_2},\ub{t_2}]$ is such that other crossing trajectories would have been possible given the assumed uncertainties, see \emph{e.g.} the red one. This proves the impossibility to both disprove this loop detection and conclude about a loop existence.}
			\label{fig:redermor:map:loopextract}
		\end{figure}

		In both Figures~\ref{fig:redermor:positions} and~\ref{fig:redermor:tplane}, the result of the degree test is displayed in green when it proves the existence of a loop and in black when nothing can be concluded. This latter case means the robot's uncertainties are too large to demonstrate that a loop has been performed or not.
		In this example, there is only one solution for which nothing can be concluded. If we have a look at Figure~\ref{fig:redermor:positions}, we can see this inconclusive case, black painted above robot's trajectory. Figure~\ref{fig:redermor:map:loopextract} provides another view of it. Looking at the reliable envelope of feasible positions pictured in gray, it could have been a loop. We know it is not the case in reality: actual trajectories are not crossing. Here, the test does not reject the feasibility of a loop, it is simply not able to conclude. 
		
		We define the actual number of loops $\actual{\lambda}$ over a mission by:
		\begin{equation}
			\actual{\lambda}=\# \big\{ \mathbf{t}\mid \actual{\mathbf{f}}(\mathbf{t})=\mathbf{0}, t_1<t_2\big\}.
		\end{equation}
		
		Now, considering uncertainties from the sensors, the theoretical number of loops proofs is given by:
		\begin{equation}
			\lambda=\# \big\{ \mathbb{T}_i \mid \forall \mathbf{f}\in[\mathbf{f}],
			\exists\mathbf{t}\in\mathbb{T}_i \mid \mathbf{f}(\mathbf{t})=\mathbf{0}\big\}.
		\end{equation}
		
		This application gives a comparison between the tests $\mathcal{T}$ and $\mathcal{N}$. Corresponding computations provide the following results:
		\begin{equation*}
			\lambda_\mathcal{N}=14 \quad\quad \lambda_\mathcal{T}=24 \quad\quad \actual{\lambda}=24
		\end{equation*}
		The blue line in Figure~\ref{fig:redermor:positions} shows the actual trajectory involves $\actual{\lambda}=24$ loops\footnote{Without considering loops in components $\Omega_i$ that intersect the boundary of $[t_0, t_f]^2$. \label{footnote:loops:included}}. On this application, no other test than the topological degree would provide better results.
			
	\subsection{The \emph{Daurade} mission}
	
\begin{figure}[!h]
			\centering
			\includegraphics[width=0.9\linewidth]{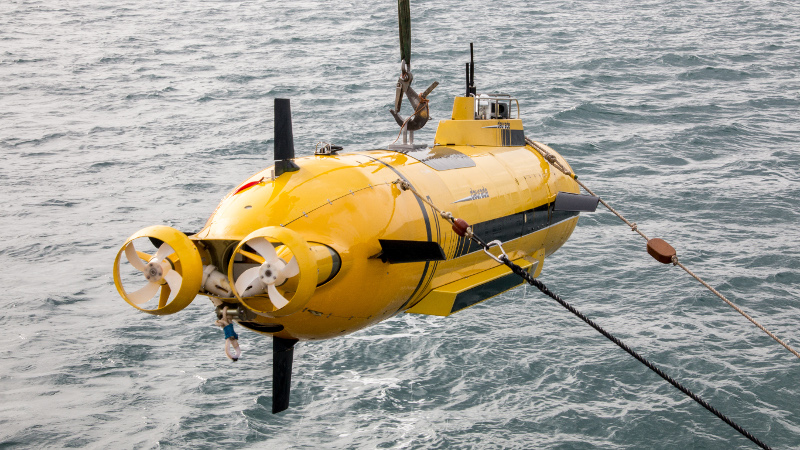}
			\caption{\emph{Daurade} AUV managed by \emph{DGA Techniques Navales Brest} and the \emph{Service Hydrographique et Oc\'{e}anographique de la Marine} (SHOM), during an experiment dedicated to this work, in the Rade de Brest, October 2015.}
			\label{fig:daurade}
		\end{figure}

\begin{figure}[!h]
			\centering
			\includegraphics[width=1.0\linewidth]{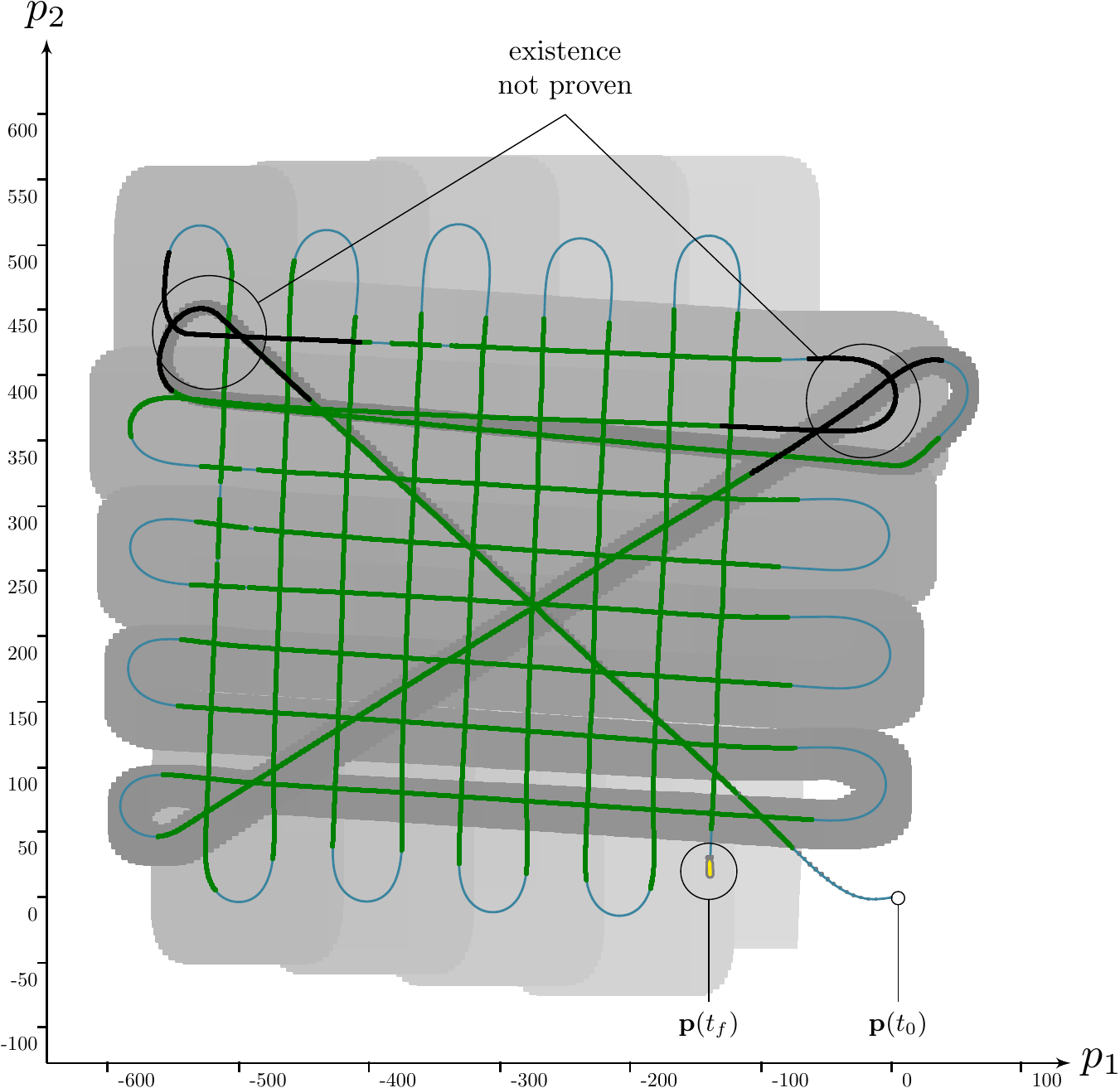}
			\caption{2D trace of \emph{Daurade} AUV. The topological test was not able to conclude for two loop detections involving a total of four actual loops. Figure~\ref{fig:daurade:map:loopextract} details one of these false detections.}
			\label{fig:daurade:map}
		\end{figure}

\begin{figure}[!h]
			\centering
			\includegraphics[width=1.0\linewidth]{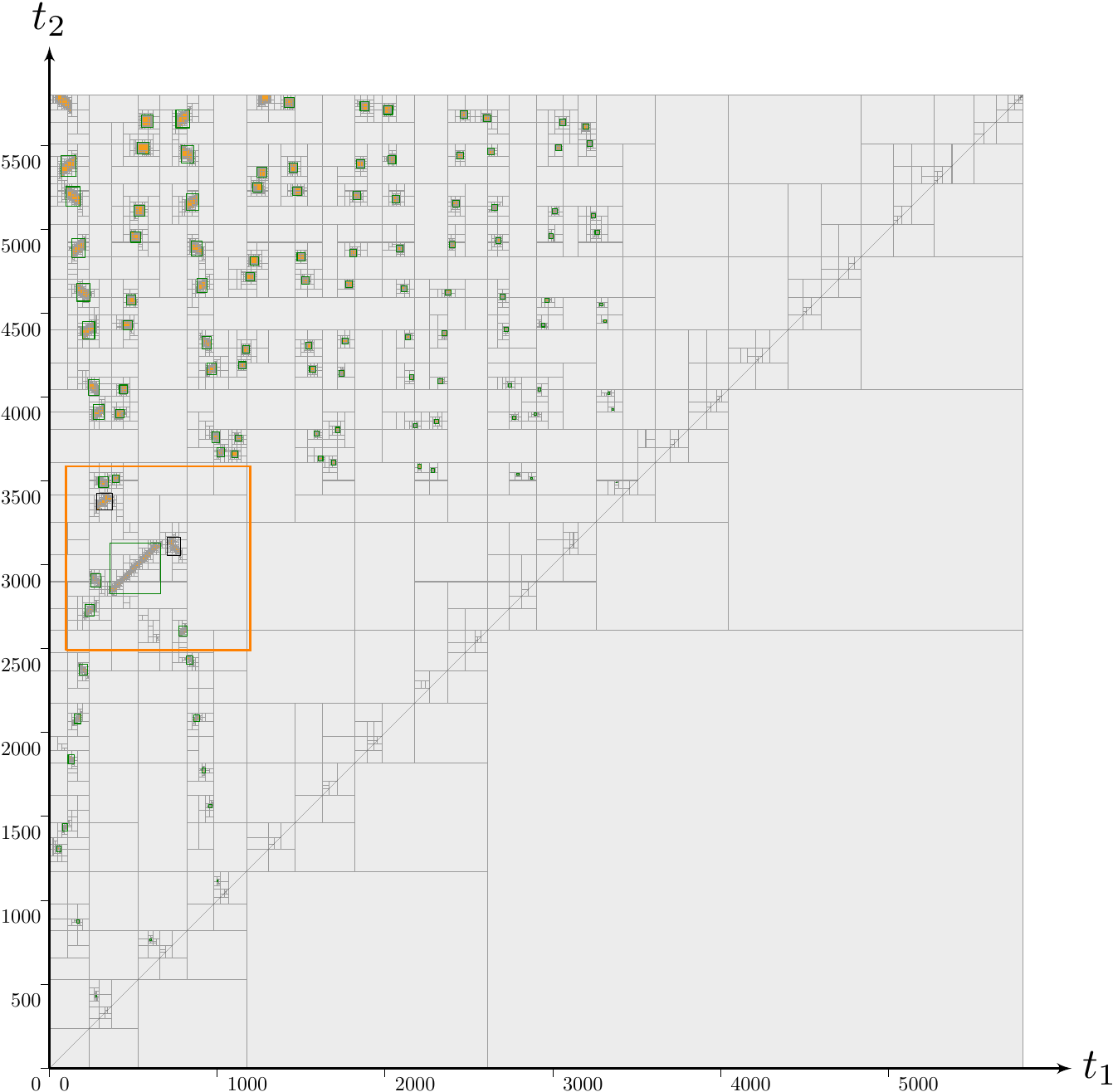}
			\caption{$t$-plane for the \emph{Daurade} experiment. The orange box is detailed in Figure~\ref{fig:daurade:tplane:zoom}.}
			\label{fig:daurade:tplane}
		\end{figure}

\begin{figure}[!h]
			\centering
			\includegraphics[width=1.0\linewidth]{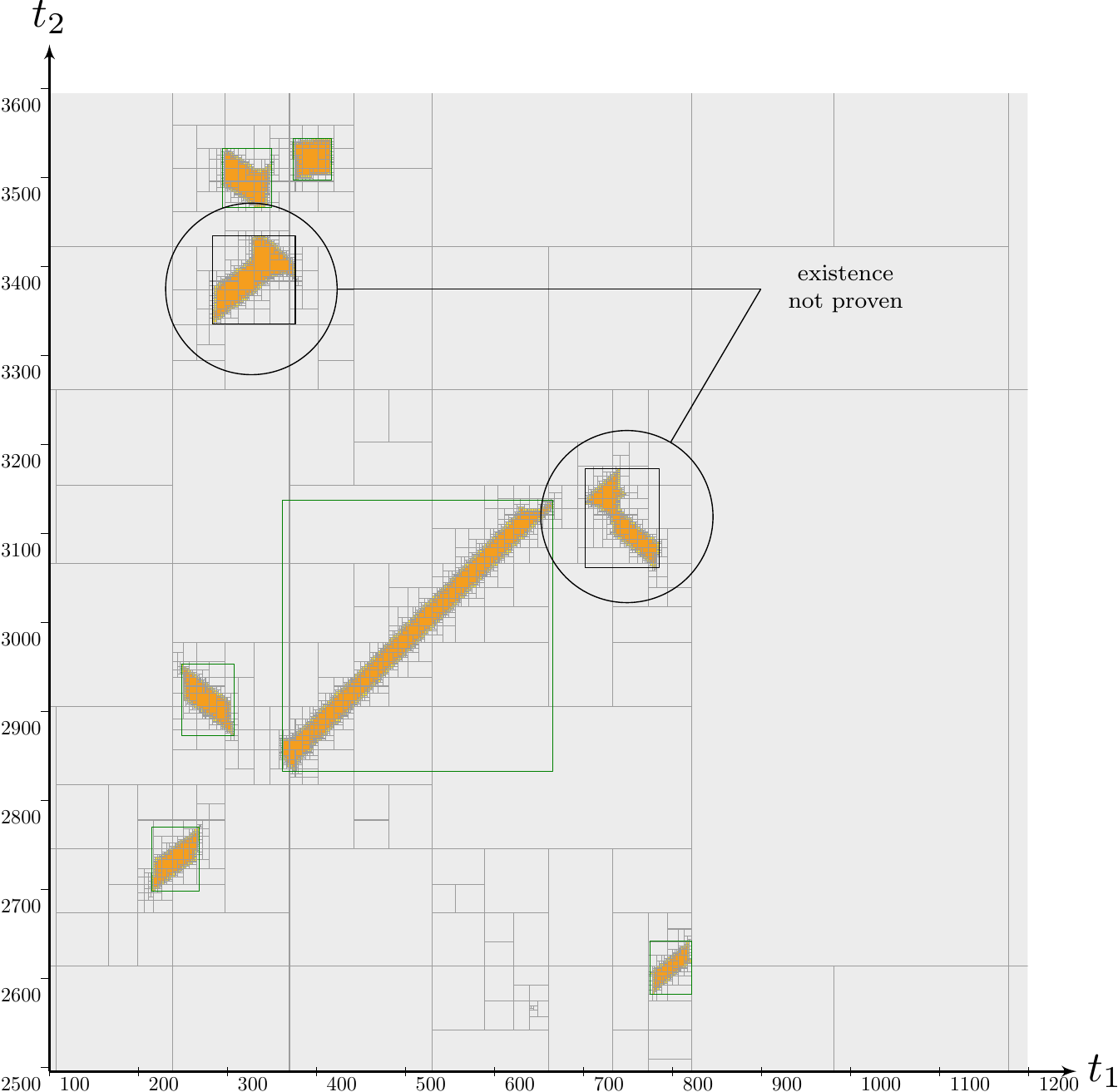}
			\caption{Zoom on $t$-plane of Figure~\ref{fig:daurade:tplane}, presenting six clusters $\Omega_i$ corresponding to loop detection sets. Two of them, black boxed, are non-conclusive cases with the topological degree test.}
			\label{fig:daurade:tplane:zoom}
		\end{figure}

		We provide a complementary example involving another AUV named \emph{Daurade}, pictured in Figure~\ref{fig:daurade}. A similar mission has been performed without surfacing during 1h40. Figure~\ref{fig:daurade:map} presents the corresponding trajectory together with its estimation and the test results. Figure~\ref{fig:daurade:tplane} and~\ref{fig:daurade:tplane:zoom} provide views of the $t$-plane.
	
		For this test case, 116 subpavings $\Omega_i$ have been computed. The test $\mathcal{T}$ proved the existence of loops in 114 of them. The uniqueness was also verified for each proof.
		Computations have been performed in less than one second on a conventional computer, which also demonstrates the relevancy of our approach for real applications.
		
\begin{figure}[!h]
			\centering
			\includegraphics[width=1.0\linewidth]{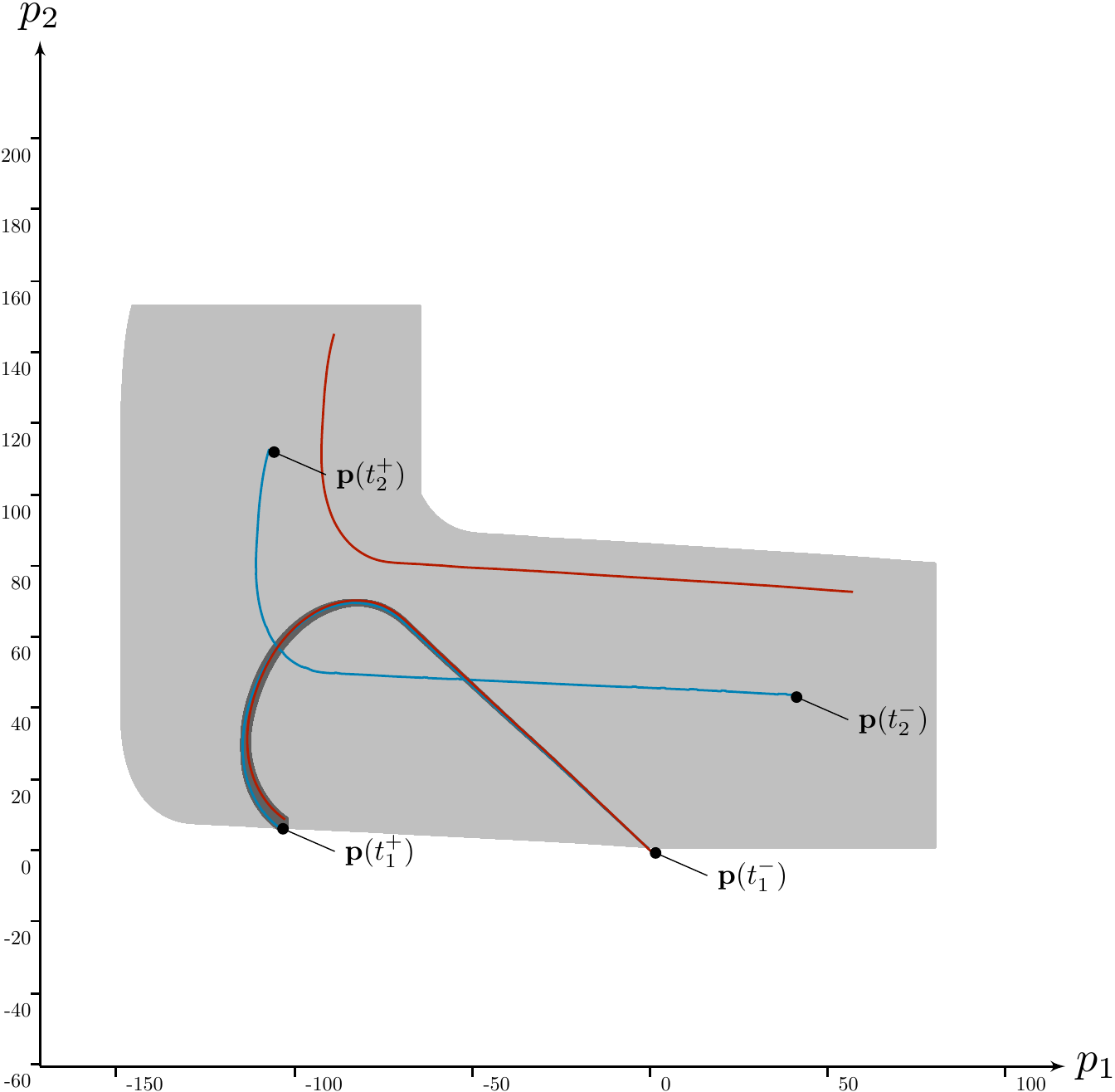}
			\caption{Independent projection of one of the two non-conclusive detection cases, as for the Redermor mission, see Figure~\ref{fig:redermor:map:loopextract}. Contrary to the previous experiment, an actual loop plotted in blue has been performed, twice. However, the red trajectory reminds that a non-crossing case is still feasible.}
			\label{fig:daurade:map:loopextract}
		\end{figure}

		The actual trajectory involved $\actual{\lambda}=118$ loops\footnoteref{footnote:loops:included} while we proved $\lambda_\mathcal{T}=114$ of them. For two loop detection sets, the algorithm did not conclude due to strong uncertainties. One of these cases is highlighted in Figure~\ref{fig:daurade:map:loopextract}.
		The next Section \ref{sec:optimal} is a discussion about the optimality of our approach. The conclusion is that in this \emph{Daurade} experiment, no more loops would have been proved by other means than the topological degree.		
		
\section{Optimality of the degree test}
\label{sec:optimal}

	In this section, we extend the aforementioned practical demonstration by a~theoretical discussion
	of the degree test and its strength.
	
\begin{figure}[!h]
		\includegraphics{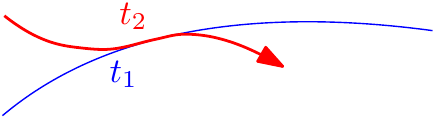}
		\caption{A ``non-transversal'' loop like this can easily be detected by the degree test, but methods requiring non-singular Jacobian matrix will fail to verify it.}
		\label{fig:singular}
	\end{figure}

	First of all, in a~situation where the interval Newton test $\mathcal{N}$ is strong enough to detect a~(unique) solution of $\actual{\mathbf{f}}(\mathbf{x})=\mathbf{0}$ in a~connected region $\Omega$, then the Jacobian matrix $\mathbf{J}_{\actual{\mathbf{f}}}$ is necessarily everywhere non-singular in $\Omega$ and the degree is either $+1$ or $-1$.
	However, the degree test does not use derivatives and can succeed even in cases where derivatives 
	are either not at hand, or when the Jacobian matrix is potentially singular. For loop detection, this includes situations such as in Figure~\ref{fig:singular}, where the self-crossing is close to parallel.
	
	Similarly, the degree test can be shown to be more powerful than other interval-based verification tests, such as Mirranda's or Borsuk's test, due to the following result~\cite[Thm 6]{quasi}:
	
	{\it 
	Whenever a~function $\actual{\mathbf{f}}$ has a \emph{robust zero} (one that cannot be removed by arbitrary small perturbations), then it can  be detected by the degree test, assuming that we have a sufficient subdivision and sufficiently precise interval-measurements.}
	
	One could still argue that such arbitrary precise interval approximations are practically not at hand. 
	Here we state another variant of the optimality of the degree, which is adapted to the setting of this paper:
	
	{\bf Proposition.}
	Let $\Omega$, $[\mathbf{f}]$, $[\mathbf{t}]_j, [\mathbf{b}]_k$ be as in~Theorem~\ref{th:topDeg:interval} and assume further that the degree $\mathrm{deg}(\actual{\mathbf{f}}, \Omega)=0$ and that the
	interior of $\Omega$ is connected.
	Then there exists a~function $\mathbf{g}\in[\mathbf{f}]$ such that
	\begin{itemize}
		\item $\mathbf{0}\notin \mathbf{g}(\Omega)$;
		\item $\mathbf{g}([\mathbf{t}]_j)\subseteq [\mathbf{f}]([\mathbf{t}]_j)$ for all $j$, and
		\item $\mathbf{g}([\mathbf{b}]_k) \subseteq [\mathbf{f}]([\mathbf{b}]_k)$ for all $k$.
	\end{itemize}
	
	In other words, whenever we detect a~zero degree on some set $\Omega$ with connected interior, then it is still possible that $\actual{\mathbf{f}}$ has no zero: indeed, the unknown function $\actual{\mathbf{f}}$ may be the function $\mathbf{g}$ from the theorem. 
	
	If we subdivided our domain more and obtained more data, our region $\Omega$ \emph{could} split into more components --- for example, $\Omega_1$ with a degree 1, and $\Omega_2$ with a degree $-1$. Each $\Omega_i$ would then provably contain a~zero. However, based only on the above interval evaluations, we cannot conclude the existence of a~zero.  In particular, for a given set of data, if we cannot conclude a~zero based on the degree test then \emph{no other test} (such as Newton) would conclude it either. 
	
	The proof of the last proposition is elementary\footnote{The main idea is to define the function $\mathbf{g}$ to be equal to $\actual{\mathbf{f}}$ on $\partial\Omega$ and, in a small enough $\epsilon$-neighborhood of the boundary, to extend it to a positive scalar multiple of $\actual{\mathbf{f}}$ such that its norm is small enough for any $x$ that is $\epsilon$-far from the boundary. This map takes $\{x:\,\,\mathrm{dist}(x,\partial\Omega)=\epsilon\}$ into a sphere of small diameter, and due to the fact that the degree is zero, can be extended to a function $\mathbf{g}: \Omega\to \R^n$ that it is still small farther from the boundary, and avoids zero.}, but requires some necessary definitions from topology, so we omit it here in order to keep the paper self-contained and readable for a~wide audience. 
	Our main message is to underline the usefulness of the degree test for zero detection of functions with bounded uncertainty, and its relevancy for loop closure proofs.

\section{Conclusion}
\label{sec:conclusion}


This paper has presented a new method to prove the existence of loops in robot trajectories. The algorithm relies on interval analysis, allowing guaranteed computations of robot trajectories by considering sensor uncertainties in a reliable way. This set-membership approach stands on measurements' bounds which allow to take conclusions by always considering worst-case possibilities. This is well suited for proof purposes and, in our case, to prove that a robot crossed its own trajectory at some point. In this approach, conclusions can be taken considering proprioceptive measurements only and no scene observation. This is helpful to solve SLAM problems as it proves a previously-visited location to be recognized.

This topic has already been the subject of previous work but the offered \emph{existence test}, relying on the Newton operator, did not give satisfactory results in some cases of undeniable looped trajectories. This was due to the use of Jacobian matrices not always invertible.
Our contribution is to propose a new test relying on the topological degree theory. The algorithm behaves better as it does not use the information of the derivatives. Besides the loop existence proof, the same tool can provide the exact number of reliable loops performed by the robot, better than the Newton test did. 
The efficiency of the new method has been demonstrated on actual experiments involving autonomous underwater robots performing several loops under the surface.

Supplementary materials are available on: \url{http://simon-rohou.fr/research/loopproof/}

{\bf Acknowledgments.}
	The work of Simon Rohou has been funded by the French \emph{Direction G\'{e}n\'{e}rale de l'Armement} (DGA) during the UK-France PhD program. The one of Peter Franek has been supported by Austrian Science Fond, M 1980.


\end{document}